\documentclass[10pt]{book}

\usepackage[table]{xcolor}
\usepackage{mathrsfs} 
\usepackage{amssymb,amsmath}
\usepackage{graphicx}
\usepackage{hyperref} 
\usepackage{xspace}
\usepackage{booktabs}
\usepackage{caption} 
\usepackage{longtable}

\usepackage{pdfpages}
\usepackage{caption}

\usepackage[T1]{fontenc}

\hypersetup{colorlinks,linkcolor={red!50!black},citecolor={blue!50!black},urlcolor={blue!80!black}}
\usepackage[top=1.2in, bottom=1.2in, left=1.4in, right=1.4in]{geometry}

\usepackage{tikz}
\usetikzlibrary{shapes,calc,positioning,automata,arrows,trees}

\newcommand\bcpen{\includegraphics[width=15pt]{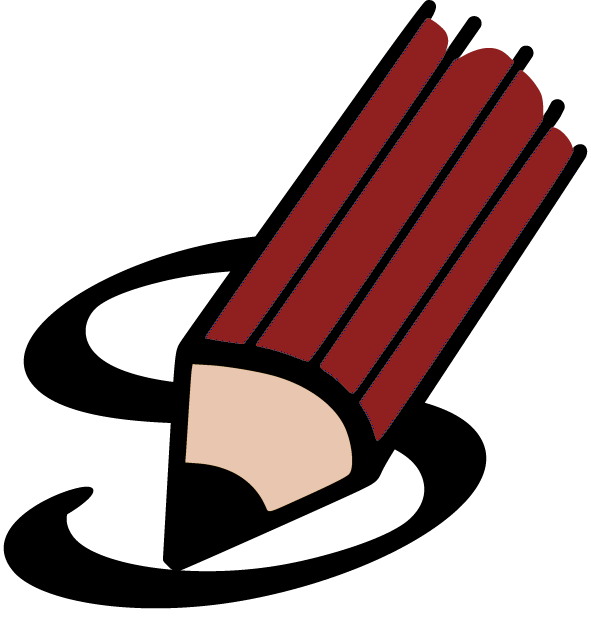}}

\usepackage{listingsutf8}
\usepackage{textcomp}
\usepackage{verbatim}
\usepackage{multirow}

\definecolor{v3lgray}{gray}{0.98}
\definecolor{v2lgray}{gray}{0.85}
\definecolor{vlgray}{gray}{0.92}
\definecolor{dgray}{rgb}{0.4,0.4,0.4}
\definecolor{dblue}{RGB}{0,0,99}
\definecolor{dred}{RGB}{175,6,54}
\definecolor{lred}{RGB}{155,6,44}
\definecolor{dgreen}{RGB}{47,135,7}
\definecolor{dviolet}{RGB}{102,0,153}
\definecolor{mblue}{RGB}{0,0,180}
\definecolor{dorange}{RGB}{204, 82, 0}

\def\ace{ACE\xspace}

\def\x3{{\rm XCSP$^3$}\xspace}
\def\jv3{{\rm JvCSP$^3$}\xspace}
\def\p3{{\rm PyCSP$^3$}\xspace}

\newcommand{\gb}[1]{{\tt #1}} 

\lstdefinelanguage{json}{
    basewidth  = {.6em,0.6em},
    basicstyle=\normalfont\ttfamily,
    breaklines=true,
    morestring=[b]',
    morestring=[b]", 
    sensitive=false,
    stringstyle=\color[rgb]{0.227,0.226,0.441}\ttfamily, 
    escapechar=!,
    showstringspaces=false,
    xleftmargin=20pt, 
    breaklines=true,basicstyle=\ttfamily\small,inputencoding=utf8/latin9,texcl
}
\lstnewenvironment{json}{\lstset{language=json}}{}

\lstdefinelanguage{mcsp}{
  keywords={forall,array,block,class,implements,model,public,slide},
  basewidth  = {.6em,0.6em},
  keywordstyle=\color{dred}\bfseries,
  ndkeywords={intension,lessThan,lessEqual,greaterEqual,greaterThan,equal,different,implication,equivalence,conjunction,disjunction,extension,regular,mdd,allDifferent,allDifferentMatrix,allEqual,ordered,increasing,decreasing,strictlyIncreasing,strictlyDecreasing,lex,lexMatrix,sum,count,atMost,atLeast,exactly,atMost1,atleast1,exactly1,element,channel,maximum,minimum,cardinality,nValues,noOverlap,cumulative,instantiation,clause,circuit,minimize,maximize},
  ndkeywordstyle=\color{mblue}\bfseries,
  identifierstyle=\color{black},
  sensitive=false,
  comment=[l]{//},
  morecomment=[s]{<!--}{-->},
  commentstyle=\color{dgreen}\ttfamily,
  stringstyle=\color{dgreen}\rmfamily, 
  morestring=[b]',
  morestring=[b]",
  escapechar=~,
  showstringspaces=false,
  classoffset=2, morekeywords={private},keywordstyle=\color{gray},
  classoffset=3, morekeywords={dom,size,when},keywordstyle=\color{dorange},
  xleftmargin=-22pt,xrightmargin=-22pt,
  breaklines=true,basicstyle=\ttfamily\footnotesize,backgroundcolor=\color{v3lgray},inputencoding=utf8/latin9,texcl
}
\lstnewenvironment{mcsp}{\lstset{language=mcsp}}{}

\definecolor{colorex}{RGB}{255,248,220} 
\definecolor{mgray}{rgb}{0.55,0.55,0.55}
\definecolor{officegreen}{rgb}{0.0, 0.5, 0.0}

\newcounter{cntPy}

\lstnewenvironment{python}{\lstset{
    inputencoding=utf8/latin1,
    language=python,
    ndkeywords={size,dom,matrix,strict,occurrences},
    ndkeywordstyle=\color{mblue}, 
    keywords=[2]{data,satisfy,minimize,maximize,solve}, 
    keywordstyle=[2]\color{dred}, 
    deletekeywords={def},
    keywords=[3]{def,If,Then,Else,Match}, 
    keywordstyle=[3]\color{dorange},
    commentstyle=\color{officegreen}, 
    showstringspaces=false,
    captionpos=t,
    basicstyle=\ttfamily\footnotesize,
    framesep=5pt,
    xleftmargin=5pt,xrightmargin=0pt,
    backgroundcolor=\color{colorex}, 
    escapechar=@
  }
}{}

\usepackage[skins,breakable,xparse]{tcolorbox}
\newcommand{\core}[1]{ 
  \medskip \begin{tcolorbox}[
    enhanced,breakable,
    boxsep=0pt,top=4pt,bottom=0pt,left=2mm,right=1mm,
    toprule=0.1mm,leftrule=0.1mm,rightrule=0.25mm,bottomrule=0.25mm,shadow={0.2mm}{-0.2mm}{0mm}{dgray},
    overlay unbroken and first={\node (logo) at ([xshift=4mm,yshift=-5mm]frame.north west) {#1}; },
    colframe=dgray,titlerule=-0.2mm,toptitle=3mm,coltitle=dred, fonttitle=\bfseries,
    lines before break=6, pad at break*=10pt
    }

\newenvironment{boxpy}
 {\stepcounter{cntPy} \core{\bcpen} , colback=colorex, title style={color=colorex}, title=~ ~ \,PyCSP$^3$ Model \thecntPy]}
 {\end{tcolorbox}}

\newenvironment{command}
  {\quote\small\verbatim}
  {\endverbatim\endquote}

  \usepackage{authblk}

\title{\textcolor{dred}{Proceedings of the \x3 Competition 2026}}
\author{Gilles Audemard \and Christophe Lecoutre \and Emmanuel Lonca}

\affil{\vspace{1cm} CRIL \\University of Artois \& CNRS \\ France}

\date{October 23, 2026}

\begin{document}
\maketitle

~ \\
~ \\

\bigskip

This document constitutes the proceedings of the \x3 Competition 2026, subsequent to those published in 2022 \cite{compet22}, 2023 \cite{compet23}, 2024 \cite{compet24} and 2025 \cite{compet25}.
The website containing all {\bf detailed results} of this international competition is available at:
\begin{quote}
  \href{https://www.cril.univ-artois.fr/XCSP26/}{https://www.cril.univ-artois.fr/XCSP26}
\end{quote}

  \bigskip
  \bigskip
\noindent The organization of this 2026 competition involved the following tasks:
\begin{itemize}
\item adjusting general details (dates, tracks, $\dots$) by G. Audemard, C. Lecoutre and E. Lonca
\item selecting instances (problems, models and data) by C. Lecoutre
\item receiving, testing and executing solvers on CRIL cluster by E. Lonca
\item validating solvers and rankings by C. Lecoutre and  E. Lonca
\item developping the 2026 website dedicated to results by G. Audemard
\end{itemize}

\bigskip\bigskip 
{\bf Important}: for reproducing the experiments and results, it is important to use the very same set of \x3 instances, as used in the competition.
These instances can be found in this \href{https://www.cril.univ-artois.fr/~lecoutre/compets/instancesXCSP26.zip}{archive}.
Some (usually minor) differences may exist when compiling the models presented in this document and those that can be found in this  \href{https://www.cril.univ-artois.fr/~lecoutre/compets/modelsXCSP26.zip}{archive}.


\tableofcontents

\chapter{About the Selection of Problems in 2026}

Remember that the complete description, {\bf Version 3.2}, of the format (\x3) used to represent combinatorial constrained problems can be found in \cite{xcsp3}.
As usual for \x3 competitions, we have limited \x3 to its kernel, called \x3-core \cite{xcsp3core}.
This means that the scope of the \x3 competition is restricted to:
\begin{itemize}
\item integer variables
\item instances of  frameworks:
  \begin{itemize}
  \item CSP (Constraint Satisfaction Problem)
  \item COP (Constraint Optimization Problem)
    \end{itemize}
\item a set of 24 popular (global) constraints for main tracks:
  \begin{itemize}
  \item generic constraints: \gb{intension} and \gb{extension} (also called \gb{table})
  \item language-based constraints: \gb{regular} and \gb{mdd}
  \item comparison constraints: \gb{allDifferent}, \gb{allDifferentList}, \gb{allEqual}, \gb{ordered}, \gb{lex} and \gb{precedence}
  \item counting/summing constraints: \gb{sum}, \gb{count}, \gb{nValues} and \gb{cardinality}
  \item connection constraints: \gb{maximum}, \gb{minimum}, \gb{element} and \gb{channel}
  \item packing/scheduling constraints: \gb{noOverlap}, \gb{cumulative}, \gb{binPacking} and \gb{knapsack}
  \item \gb{circuit}, \gb{instantiation} (and \gb{slide})
  \end{itemize}
  and a smaller set of constraints for mini tracks.
\end{itemize}

\medskip
For the 2026 competition, 31 problems have been selected.
They are succinctly presented in Table \ref{fig:problems}.
For each problem, the type of the involved (global) constraints is indicated.
At this point, do note that making a good selection of problems/instances is a difficult task.
In our opinion, important criteria for a good selection are:
\begin{itemize}
\item the novelty of problems, avoiding constraint solvers to overfit already published problems;
\item the diversity of constraints, trying to represent all of the most popular constraints (those from \x3-core) while paying attention to not over-representing some of them;
\item the scaling up of problems.
\end{itemize}

\begin{table}
  \begin{tabular}{p{5.5cm}p{0.2cm}p{7.5cm}} 
    \toprule
CSP Problems & & Global Constraints\\
    \midrule
\rowcolor{vlgray}{}  AllDifferentPathological & & \gb{allDifferent}  \\
    Antichain1 &   &   \gb{count}, \gb{lex} \\
\rowcolor{vlgray}{}  Antichain2   &  & \gb{element}, \gb{lex}  \\
    BIBD2 & &  \gb{cardinality}, \gb{lex}, \gb{nValues}  \\
\rowcolor{vlgray}{}   Collatz & &  \gb{allDifferent} \\
    ColoredGrid & & \gb{allDifferent}, \gb{lex} \\
\rowcolor{vlgray}{}  Dominoes & & \gb{allDifferent}, \gb{table} \\
GeneralizedPigeonHole & &  \gb{allDifferent} \\
\rowcolor{vlgray}{} HIPP & & \gb{element} \\
Inglenook & & \gb{table} \\
\rowcolor{vlgray}{} KillerSudoku & & \gb{allDifferent}, \gb{sum} \\
MagicHexagon &  & \gb{allDifferent}, \gb{sum} \\
\rowcolor{vlgray}{} PerfectRythmicTiling1 & & \gb{allDifferent}, \gb{cardinality}, \gb{element}, \gb{table} \\
PerfectRythmicTiling2 &  & \gb{allDifferent}, \gb{table} \\
\rowcolor{vlgray}{} RectPacking & & \gb{noOverlap} \\
Steiner3 &  & \gb{table} \\
& & \\
\midrule
COP Problems & & Global Constraints \\ 
\midrule
\rowcolor{vlgray}{} BACPj &   & \gb{binPacking}, \gb{sum} \\
 Clique &   &  \gb{sum} \\
\rowcolor{vlgray}{} CrosswordDesign &   & \gb{sum}, \gb{table} \\
EternityApprox &  & \gb{allDifferent}, \gb{cardinality}, \gb{sum}, \gb{table} \\
\rowcolor{vlgray}{} ExaminationTimetabling1 &   & \gb{element}, \gb{sum}, \gb{table} \\
ExaminationTimetabling3 &  & \gb{binPacking}, \gb{element}, \gb{sum}, \gb{table} \\
\rowcolor{vlgray}{} Feedback &  & \gb{allDifferent}, \gb{sum} \\
GolombRuler &  & \gb{allDifferent}  \\
\rowcolor{vlgray}{} MaxAgreemeentForest2 &  & \gb{allDifferent}, \gb{allEqual}, \gb{element}, \gb{sum}, \gb{table} \\
MicroplateSwapping &   & \gb{count}, \gb{element}, \gb{sum}, \gb{table}  \\
\rowcolor{vlgray}{} NonDominatingQueens &  & \gb{count}, \gb{lex}, \gb{sum}, \gb{table} \\
SatelliteObservations &  & \gb{allDifferent}, \gb{sum}, \gb{table} \\
\rowcolor{vlgray}{} Sonet2 &  & \gb{count}, \gb{lex}, \gb{sum} \\
SumSquares &  &  \gb{sum} \\
\rowcolor{vlgray}{} TelecomScheduling &   & \gb{count}, \gb{element}, \gb{maximum}, \gb{precedence}, \gb{sum}, \gb{table} \\
\bottomrule
  \end{tabular}
  \caption{Selected Problems for the main tracks of the 2026 Competition. When \gb{table} is followed by ($*$), it means that starred tables are involved.}\label{fig:problems}
\end{table}

\paragraph{Novelty.} Almost all problems are new in 2026, with models directly written in \p3.



\paragraph{Scaling up.} It is always interesting to see how constraint solvers behave when the instances of a problem become harder and harder.
This is what we call the scaling behavior of solvers.
For most of the problems in the 2026 competition, we have selected series of instances with regular increasing difficulty.
It is important to note that assessing the difficulty of instances was mainly determined with \ace \cite{ace}, which is the reason why \ace is declared to be off-competition (due to this strong bias).

\paragraph{Selection.} This year, the selection of problems and instances has been performed by Christophe Lecoutre.
As a consequence, the solver \ace was labeled off-competition.


\bigskip

\chapter{Problems and Models}

In the next sections, you will find all models used for generating the \x3 instances of the 2026 competition (for main CSP and COP tracks).
All models are written in \p3 \cite{pycsp3}, Version 2.6 (c) (released in March 1, 2026); see \href{https://pycsp.org}{https://pycsp.org}.

\section{CSP}

\subsection{AllDifferentPathological}

\paragraph{Description.}
This problem is used in the paper by J.-F. Puget in \cite{P_fast}.
It is introduced to show the worst-case behavior (pathological behavior) of some filtering algorithms.

\paragraph{Data.}
An integer is required to specify a specific instance: the order $n$ of the problem instance. 
The values used for generating the 2026 competition instances are:
\begin{quote}
10000, 20000, 30000, 40000, 50000, 60000, 70000, 80000, 90000, 100000 
\end{quote}

\paragraph{Model.}
The \p3 model, in a file `AllDifferentPathological.py', used for the competition is: 

\begin{boxpy}\begin{python}
@\imp@

n = data 

x = VarArray(
      size=2 * n + 1,
      dom=lambda i: range(i - n, 1) if i <= n else range(0, i - n + 1)
   )

satisfy(
   AllDifferent(x)
)
\end{python}\end{boxpy}

This model involves a single array of variables, the domains of which are specific.
A series of 10 instances has been selected for the competition.
For generating an \x3 instance (file), you can execute for example:
\begin{command}
  python AllDifferentPathological.py -data=10000
\end{command}

\subsection{Antichain1}

\paragraph{Description.}

The antichain problem is to find a set of multisets under some conditions.
Here, we must find $n$ multisets, drawn from $\{1, \dots, d\}$, where for each pair of multisets $v_1$ and $v_2$, neither of $v_1 \leq v_2$ or $v_1 \leq v_2$ hold.

\paragraph{Data.}

Two integers are required to specify a specific instance: the number of (multi-)sets and the number of values.
The pairs of integer values used for generating the 2026 competition instances are:
\begin{quote}
(20, 6), (21, 6), (35, 7), (36, 7), (70, 8), (71, 8), (126, 9), (127, 9), (252, 10), (253,10) 
\end{quote}

\paragraph{Model.}
The \p3 model, in a file `Antichain1.py', used for the competition is: 

\begin{boxpy}\begin{python}
@\imp@

nSets, nVals = data 

# x[i][j] is 1 if value j is present in the ith set
x = VarArray(size=[nSets, nVals], dom={0, 1})

satisfy(
   # ensuring no set dominates another one
   [
      (
         Exist(x[i1][j] < x[i2][j] for j in range(nVals)),
         Exist(x[i1][j] > x[i2][j] for j in range(nVals))
      ) for i1, i2 in combinations(nSets, 2)
   ],

   # tag(symmetry-breaking)
   LexIncreasing(x)
)
\end{python}\end{boxpy}

This model involves a single array of variables, some global constraints of type \gb{count} and \gb{lex}.
A series of 10 instances has been selected for the competition.
For generating an \x3 instance (file), you can execute for example:
\begin{command}
  python Antichain1.py -data=[20,6] 
\end{command}

\subsection{Antichain2}

\paragraph{Description.}

The antichain problem is to find a set of multisets under some conditions.
Here, we must find $n$ multisets, drawn from $\{1, \dots, d\}$, where for each pair of multisets $v_1$ and $v_2$, neither of $v_1 \leq v_2$ or $v_1 \leq v_2$ hold.

\paragraph{Data.}

Three integers are required to specify a specific instance.
The values $(n,d,p)$ used for generating the 2026 competition instances are:
\begin{quote}
  (19, 4, 3), (20, 4, 3), (51, 5, 3), (52, 5, 3), (141, 6, 3), (142, 6, 3), (37, 3, 7), \\
  (38, 3, 7), (44, 4, 4), (45, 4, 4), (85, 4, 5), (86, 4, 5), (100, 3, 10), (120, 3, 10)
\end{quote}

\paragraph{Model.}
The \p3 model, in a file `Antichain2.py', used for the competition is: 

\begin{boxpy}\begin{python}
@\imp@

n, d, p = data 

# x[i][j] is the number of occurrences of  value j in the ith set
x = VarArray(size=[n, d], dom=range(p))

xmin = VarArray(size=[n, n], dom=lambda i, j: range(d) if i < j else None)

xmax = VarArray(size=[n, n], dom=lambda i, j: range(d) if i < j else None)

satisfy(
   # ensuring no set dominates another one
   [
      (
         x[i1][xmin[i1][i2]] < x[i2][xmin[i1][i2]],
         x[i1][xmax[i1][i2]] > x[i2][xmax[i1][i2]],
      ) for i1, i2 in combinations(n, 2)
   ],

   # tag(symmetry-breaking)
   LexIncreasing(x)
)
\end{python}\end{boxpy}

This model involves two arrays of variables, and some global constraints of type \gb{element} and \gb{lex}.
A series of 14 instances has been selected for the competition.
For generating an \x3 instance (file), you can execute for example:
\begin{command}
  python Antichain2.py -data=[19,4,3]
\end{command}

\subsection{BIBD2}

\paragraph{Description.}
A BIBD (Balanced Incomplete Block Design) is defined as an arrangement of $v$ distinct objects into $b$ blocks such that each block contains exactly $k$ distinct objects,
each object occurs in exactly $r$ different blocks, and every two distinct objects occur together in exactly $ld$ blocks.
Another way of defining a BIBD is in terms of its incidence matrix, which is a $v$ by $b$ binary matrix with exactly $r$ ones per row, $k$ ones per column,
and with a scalar product of $ld$ between any pair of distinct rows.
A BIBD is therefore specified by its parameters $(v,b,r,k,ld)$.

Here, this is a model (not necessarily efficient) used as a benchmark for the 2026 \x3 Competition.

\paragraph{Data.}

Three integers are required to specify a specific instance.
The values $(v,k,ld)$ used for generating the 2026 competition instances are:
\begin{quote}
  (8, 4, 3), (6, 3, 6), (11, 5, 2), (10, 4, 2), (7, 3, 3), (6, 3, 8),
  (9, 4, 3), (7, 3, 4), (6, 3, 10), (9, 3, 2), (16, 6, 2), (19, 9, 4)
\end{quote}

\paragraph{Model.}
The \p3 model, in a file `BIBD2.py', used for the competition is: 

\begin{boxpy}\begin{python}
@\imp@

v, b, r, k, ld = data 
b = (ld * v * (v - 1)) // (k * (k - 1)) if b == 0 else b  # when 0, b is computed
r = (ld * (v - 1)) // (k - 1) if r == 0 else r  # when 0, r is computed


# x[i][j] is the value of the matrix at row i and column j
x = VarArray(size=[v, r], dom=range(b))

satisfy(
   # ordering values on each row
   [Increasing(x[i], strict=True) for i in range(v)],

   # constraints on columns
   Cardinality(within=x, occurrences={j: k for j in range(b)}),

   # scalar constraints with respect to lambda
   [NValues(x[i1] + x[i2]) == 2 * r - ld for i1, i2 in combinations(v, 2)],

   # Increasingly ordering rows tag(symmetry-breaking)
   [x[i] <= x[i + 1] for i in range(v - 1)]
)

\end{python}\end{boxpy}

This model involves a single array of variables, and some global constraints of type \gb{cardinality}, \gb{nValues} and \gb{lex}.
A series of 12 instances has been selected for the competition.
For generating an \x3 instance (file), you can execute for example:
\begin{command}
  python BIBD2.py -data=[8,4,3]
\end{command}

\subsection{Collatz}

\paragraph{Description.}
The following iterative sequence is defined for the set of positive integers:
\begin{itemize}
\item $n \rightarrow n/2$ if n is even
\item $n \rightarrow 3n + 1$ if n is odd
\end{itemize}

Using the rule above and choosing a starting value (while limiting the maximal value), we can compute the sequence of values.
Although it has not been proved yet (Collatz Problem), it is thought that all starting numbers finish at 1.
See \href{https://projecteuler.net/problem=14}{https://projecteuler.net/problem=14}.

\paragraph{Data.}

Two integers are required to specify a specific instance.
The values $(h,d)$ used for generating the 2026 competition instances are:
\begin{quote}
(100, 10000), (150, 10000), (200, 10000), (100, 25000), (175, 25000), (200, 25000),
(100, 50000), (190, 50000), (200, 50000), (200, 100000), (250, 100000), (300, 100000)
\end{quote}

\paragraph{Model.}
The \p3 model, in a file `Collatz.py', used for the competition is: 

\begin{boxpy}\begin{python}
@\imp@

horizon, maxValue = data 

# x[i] is the ith value of the Collatz sequence
x = VarArray(size=horizon, dom=lambda i: {1} if i == horizon - 1 else range(2, maxValue + 1))

satisfy(
   # computing the Collatz sequence
   [
      If(
         x[i] % 2 == 0,
         Then=x[i + 1] == x[i] // 2,
         Else=x[i + 1] == 3 * x[i] + 1
      ) for i in range(horizon - 1)
   ],

   AllDifferent(x)
)
\end{python}\end{boxpy}

This model involves a single array of variables, some arithmetic constraints and one constraint \gb{allDifferent}.
A series of 14 instances has been selected for the competition.
For generating an \x3 instance (file), you can execute for example:
\begin{command}
  python Collatz.py -data=[100,10000]
\end{command}

\subsection{Colored Grid}

\paragraph{Description.}

See Allan Gottlieb's Home Page (\href{https://cs.nyu.edu/~gottlieb/tr/overflow/2010-jul-aug-2-1.pdf}{part1} and \href{https://cs.nyu.edu/~gottlieb/tr/overflow/2010-jul-aug-2-2.pdf}{part2}).
The objective is to fill in an $n \times n$ grid with digits $\{1,2,\dots,n\}$ so that no digit repeats horizontally, vertically or in a non-wrapping diagonal (with as few holes as possible).

\paragraph{Data.}

One integer is required to specify a specific instance.
The values used for generating the 2026 competition instances are:
\begin{quote}
6, 7, 8, 9, 10, 11, 12, 13, 14, 17, 19, 23
\end{quote}

\paragraph{Model.}
The \p3 model, in a file `ColoredGrid.py', used for the competition is: 

\begin{boxpy}\begin{python}
@\imp@

n = data or 8

symmetries = [f.apply_on(n) for f in TypeSquareSymmetry]

# x[i][j] is the value in the cell at row i and column j
x = VarArray(size=[n, n], dom=range(n))

satisfy(
   # ensuring different values in each row and in each column
   AllDifferent(x, matrix=True),

   # ensuring different values ine ach diagonal
   [AllDifferent(diag) for diag in diagonals_up(x) + diagonals_down(x)],
   
   # tag(symmetry-breaking)
   [x <= x[symmetry] for symmetry in symmetries]
)
\end{python}\end{boxpy}

This model involves a single array of variables, and some global constraints of type \gb{allDifferent} and \gb{lex}.
A series of 12 instances has been selected for the competition.
For generating an \x3 instance (file), you can execute for example:
\begin{command}
  python ColoredGrid.py -data=8
\end{command}

\subsection{Dominoes}

\paragraph{Description.}

You are given a grid of size $n \times m$ containing numbers being parts of dominoes.
For example, for $n=7$ and $m=8$, the grid contains all dominoes from 0-0 to 6-6.
One has to find the position (and rotation) of each domino; see e.g., \cite{S_teaching,S_dominoes}.

\paragraph{Data.}

As an illustration of data specifying an instance of this problem, we have:
\begin{json}
{
  "grid": [
    [0, 5, 2, 2, 5, 4, 6, 5],
    [3, 6, 2, 2, 4, 4, 4, 1],
    [3, 6, 1, 2, 3, 4, 6, 1],
    [0, 1, 4, 3, 0, 2, 2, 1],
    [3, 5, 3, 0, 3, 1, 5, 6],
    [6, 4, 0, 3, 6, 0, 4, 1],
    [1, 6, 0, 0, 2, 5, 5, 5]
  ]
}
\end{json}

\paragraph{Model.}

The variant `table' of the \p3 model, in a file `Dominoes.py', used for the \x3 competition is:

\begin{boxpy}\begin{python}
@\imp@

grid = data 

nRows, nCols, nValues = len(grid), len(grid[0]), len(grid)
dominoes = [(i, j) for i in range(nValues) for j in range(i, nValues)]  
indexes = range(nRows * nCols)  # indexes of cells

P = [[i * nCols + j for i in range(nRows) for j in range(nCols) if grid[i][j] == value]
       for value in range(nValues)]  # possible positions

# x[i][j] is the index (position) in the grid of the value i of the domino i-j
x = VarArray(size=[nValues, nValues], dom=lambda i, j: indexes if i <= j else None)

# y[i][j] is the index (position) in the grid of the value j of the domino i-j
y = VarArray(size=[nValues, nValues], dom=lambda i, j: indexes if i <= j else None)

satisfy(
   # each part of each domino in a different cell
   AllDifferent(x + y),

   # unary constraints
   [
      (
         x[i][j] in P[i],
         y[i][j] in P[j]
      ) for i, j in dominoes
   ]
)

T = [(k1, k2) for k1 in indexes for k2 in indexes
        if abs(k1 - k2) == nCols or abs(k1 - k2) == 1 and k1 // nCols == k2 // nCols]

satisfy(
   # adjacency constraints
   (x[i][j], y[i][j]) in T for i, j in dominoes
)
\end{python}\end{boxpy}

This model involves two arrays of variables, and \gb{allDifferent} and \gb{table} constraints.
A series of 12 instances has been selected for the competition.
For generating an \x3 instance (file), you can execute for example:
\begin{command}
  python Dominoes.py -data=grid01.json
\end{command}

\subsection{Generalized Pigeon Hole}

\paragraph{Description.}
The problem is to find assignments to a two-dimensional array of variables, where every pair of rows in the array must be unequal.
See \cite{JMNP_logical}.

\paragraph{Data.}

Three integers are required to specify a specific instance.
Indeed, the parameters for this problem are the number of rows $n$, the number of columns $m$, and the domain size $d$.
The values $(n,m,d)$ used for generating the 2026 competition instances are:
\begin{quote}
  (32, 5, 2), (33, 5, 2), (64, 6, 2), (65, 6, 2), (128, 7, 2), (256, 8, 2), (512, 9, 2),\\
  (1024, 10, 2), (243, 5, 3), (729, 6, 3), (2187, 7, 3), (1024, 5, 4), (4096, 6, 4),\\
  (3125, 5, 5), (7796, 5, 6) 
\end{quote}

\paragraph{Model.}
The \p3 model, in a file `GeneralizedPigeonHole.py', used for the competition is: 

\begin{boxpy}\begin{python}
@\imp@

n, m, d = data 

# x[i][j] is the value of the jth element of the ith vector
x = VarArray(size=[n, m], dom=range(d))

satisfy(
   # ensuring different vectors
   AllDifferentList(x)
)        
\end{python}\end{boxpy}

This model involves a single array of variables, and a global constraint of type \gb{allDifferentList}.
A series of 13 instances has been selected for the competition.
For generating an \x3 instance (file), you can execute for example:
\begin{command}
  python GeneralizedPigeonHole.py -data=[32,5,2]
\end{command}

\subsection{HIPP}

\paragraph{Description.}

Haplotype inference is the problem of determining the haplotypes that form a given set of genotypes.
Different pairs of haplotypes may form the same genotype, and this ambiguity makes it difficult to find the `correct' haplotypes that explain the given genotypes.
Haplotype Inference by Pure Parsimony (HIPP) is a slight modification of the haplotype inference problem, which is to infer a minimal set of haplotypes that explain the given genotypes.
The decision version of HIPP (i.e., deciding that a set of $k$ haplotypes that explain the given genotypes exists) is NP-hard.
See \cite{G_haplotype,LPR_haplotyping,CET_comparing}.

\paragraph{Data.}

As an illustration of data specifying an instance of this problem, we have:
\begin{json}
{
  "n":10,
  "m":41,
  "k":16,
  "g":[
    [0,0,0,0,0,0,0,0,0,0,0,0,0,2,0,2,0,0,0,0,0,0,0,0,0,0,0,0,1,0,1,0,0,1,...,1,0],
    [0,0,0,0,2,2,2,2,0,2,0,1,0,1,0,1,1,0,1,0,0,0,0,0,1,1,1,0,0,0,0,0,1,1,...,1,0],
    [0,0,0,0,1,0,0,0,1,0,0,0,0,2,0,2,0,0,0,0,1,0,0,0,0,0,0,1,0,0,0,0,0,0,...,2,0],
    [0,0,0,0,1,0,0,1,0,0,0,1,0,0,0,1,0,1,1,1,1,1,0,1,1,1,0,0,1,1,1,1,1,1,...,0,1],
    [0,0,0,0,1,1,1,1,0,0,0,1,0,1,0,1,1,0,1,0,0,0,0,0,1,1,1,0,1,0,0,1,1,1,...,0,0],
    [0,0,0,0,1,1,1,1,1,0,0,1,0,1,1,1,0,1,1,0,1,1,0,0,0,0,0,0,1,1,1,1,1,0,...,2,0],
    [2,2,2,0,0,0,0,2,0,0,0,1,1,1,1,0,1,1,2,1,1,2,1,2,2,1,1,1,1,1,1,2,1,2,...,0,1],
    [1,0,1,1,1,1,0,1,0,1,0,1,0,1,0,1,1,0,1,0,0,0,0,0,1,1,1,0,0,0,0,1,1,1,...,1,0],
    [1,1,2,0,2,0,0,1,1,0,0,2,1,0,1,0,0,1,2,1,1,1,0,1,2,0,0,1,2,2,2,2,2,2,...,0,0],
    [1,1,1,0,1,0,0,1,0,0,0,0,0,1,0,0,1,0,1,0,0,0,0,0,1,1,0,1,1,1,1,2,1,0,...,2,0]
  ]
}
\end{json}

Data files can be found at \href{https://people.sabanciuniv.edu/esraerdem/benchmarks}{https://people.sabanciuniv.edu/esraerdem/benchmarks/}.

\paragraph{Model.}

The \p3 model, in a file `HIPP.py', used for the \x3 competition is:

\begin{boxpy}\begin{python}
@\imp@

n, m, k, g = data

N, M = range(n), range(m)

# h[q][j] is 1 iff the jth element is present in the qth haplotype
h = VarArray(size=[k, m], dom={0, 1})

# s[i] is the mapping of every genotype i to two haplotypes
s = VarArray(size=[n, 2], dom=range(k))

if variant('brown'):

   satisfy(
      [h[s[i][0]][j] != h[s[i][1]][j] for i in N for j in M if g[i][j] == 1],

      [h[s[i][0]][j] + h[s[i][1]][j] == g[i][j] for i in N for j in M if g[i][j] != 1]
   )

else:

   satisfy(
      # tag(redundant)
      [s[i][0] != s[i][1] for i in N if any(g[i][j] == 2 for j in M)],  

      [h[s[i][0]][j] != h[s[i][1]][j] for i in N for j in M if g[i][j] == 2],

      [
         [
            h[s[i][0]][j] == g[i][j],
            h[s[i][1]][j] == g[i][j]
         ] for i in N for j in M if g[i][j] != 2
      ]
   )
\end{python}\end{boxpy}

This model involves two arrays of variables, and global constraints of type \gb{element}.
A series of $2 \times 8 = 16$ instances has been selected for the competition (here, we have a main variant and a variant `brown').
For generating an \x3 instance (file), you can execute for example:
\begin{command}
  python HIPP.py -data=[I0-gus.json,k=15]
  python HIPP.py -data=[I0-bro.json,k=15] -variant=brown
\end{command}

\subsection{Inglenook}

\paragraph{Description.}

From \cite{SQ_inglenook}:
\begin{quote}
  The Inglenook shunting puzzle is an example of a railway based planning problem.
  Given an initial state of wagons placed in the layout, the aim is to achieve a given target state in as few moves as possible.
  Each move is an operation involving a locomotive (also called shunter) moving wagons between locations of a limited capacity,
  coupling and uncoupling wagons as required.
  This is a very popular problem for railway enthusiasts, with many articles and videos describing how to build such a puzzle layout.
\end{quote}

Here, this the state-based model as proposed by H. Simonis and L. Quesada.

\paragraph{Data.}

As an illustration of data specifying an instance of this problem, we have:
\begin{quote}
([1, 2, 3, 4, 0], [0, 5, 0], 3),
\end{quote}

\paragraph{Model.}

The \p3 model, in a file `Inglenook.py', used for the \x3 competition is:

\begin{boxpy}\begin{python}
@\imp@

startA, startB, nMoves = data

nWagons, nStates, nActions, start_state, final_states, T = build_structures(startA, startB)
# code of build_structures not shown here

# x{i] is the state after the ith move
x = VarArray(size=nMoves + 1, dom=range(nStates))

satisfy(
   # setting the initial state
   x[0] == start_state,

   # ensuring a valid final state
   x[-1] in final_states
)


satisfy(
   # ensuring valid transitions of states
   Table(
      scope=(x[i], x[i + 1]),
      supports=T
   ) for i in range(nMoves)
)
\end{python}\end{boxpy}

This model involves a single array of variables, and global constraints of type \gb{table}.
A series of 16 instances has been selected for the competition.
For generating an \x3 instance (file), you can execute for example:
\begin{command}
  python Inglenook.py -data=[startA=[1,2,3,4,0],startB=[0,5,0],nMoves=3]  
\end{command}

\subsection{Killer Sudoku}

This is Problem \href{https://www.csplib.org/Problems/prob057/}{057} at CSPLib.

\paragraph{Description.}

This is the Sudoku puzzle with a set of cages.
A cage is a set of contiguous cells with numbers in them that have to add up to a total.
Also, the cells in a cage cannot contain the same number more than once. 
The cages do not overlap, and they cover all cells.

\paragraph{Data.}

As an illustration of data specifying an instance of this problem, we have:

\begin{json}
{
  "n": 16,
  "cages": [
    [20, 0, 16],
    [37, 80, 81, 65, 49],
    ...,
    [44, 238, 239, 255, 254]
  ]
}
\end{json}

\paragraph{Model.}

The \p3 model, in a file `KillerSudoku.py', used for the \x3 competition is:

\begin{boxpy}\begin{python}
@\imp@

n, raw_cages = data  # n (order of the grid) is typically 9
base = int(sqrt(n))


# x[i][j] is the value in cell at row i and col j.
x = VarArray(size=[n, n], dom=range(1, n + 1))

cages = [(cage[0], [x[v // n][v % n] for v in cage[1:]]) for cage in raw_cages]

satisfy(
   # imposing distinct values on each row and each column
   AllDifferent(x, matrix=True),

   # imposing distinct values on each block  tag(blocks)
   [AllDifferent(x[i:i + base, j:j + base]) for i in range(0, n, base)
       for j in range(0, n, base)],

   # each cage must add up to the right value
   [Sum(cells) == v for v, cells in cages],

   # each cage must contain different values
   [AllDifferent(cells) for _, cells in cages]
)
\end{python}\end{boxpy}

This model involves a single array of variables, and global constraints of type \gb{allDifferent} and \gb{sum}.
A series of 16 instances has been selected for the competition.
For generating an \x3 instance (file), you can execute for example:
\begin{command}
  python KillerSudoku.py -data=<datafile.json>
  python KillerSudoku.py -data=<datafile.txt> -parser=KillerSudoku_ParserE.py
\end{command}

\subsection{Magic Hexagon}

This is Problem \href{http://www.csplib.org/Problems/prob023}{023} on CSPLib.

\paragraph{Description.}

A magic hexagon  consists of the numbers 1 to 19 arranged in a hexagonal pattern.

\paragraph{Data.}

We just need to specify two integers $n$ and $s$.

\paragraph{Model.}

The \p3 model, in a file `MagicHexagon.py', used for the \x3 competition is:

\begin{boxpy}\begin{python}
@\imp@

n, s = data 

domain = range(s, s + 3 * n * n - 3 * n + 1)
magic = sum(domain) // (2 * n - 1)

d = n + n - 1  # longest diameter
D = range(d)

def scope(i, right):
    v1 = max(0, d // 2 - i if right else i - d // 2)
    v2 = d // 2 - v1
    return [x[j + v1][i - max(0, v2 - j if right else j - v2)]
               for j in range(d - abs(d // 2 - i))]


# x represents the hexagon; on row x[i], only the first n - |n/2 - i| cells are useful
x = VarArray(size=[d, d], dom=lambda i, j: domain if j < d - abs(d // 2 - i) else None)

satisfy(
   # putting each value exactly once
   AllDifferent(x),

   # all rows sum to the magic value
   [Sum(x[i]) == magic for i in D],

   # all right-sloping diagonals sum to the magic value
   [Sum(scope(i, True)) == magic for i in D],

   # all left-sloping diagonals sum to the magic value
   [Sum(scope(i, False)) == magic for i in D],

   # tag(symmetry-breaking)
   [
      x[0][0] < x[0][n - 1],
      x[0][0] < x[n - 1][d - 1],
      x[0][0] < x[d - 1][n - 1],
      x[0][0] < x[d - 1][0],
      x[0][0] < x[n - 1][0],
      x[0][n - 1] < x[n - 1][0]
   ]
)
\end{python}\end{boxpy}

This model involves a single array of variables, and global constraints of type \gb{allDifferent} and \gb{sum}.
A series of 11 instances has been selected for the competition.
For generating an \x3 instance (file), you can execute for example:
\begin{command}
  python MagicHexagon.py -data=[number,number]
\end{command}

\subsection{PerfectRythmingTiling}

\paragraph{Description.}

A perfect rhythmic tiling of order $k$ and dimension $n$ is a finite sequence of $k*n$ integer terms such that:
\begin{itemize}
\item every integer from 0 to $n-1$, inclusive, is equal to exactly $k$ terms of the sequence,
\item the ranks of these $k$ equal terms form an arithmetic progression,
\item and the common ratios of these progressions are all different.
\end{itemize}
It can also be defined as a partition of the first $k*n$ natural numbers into $n$ sets of $k$ elements that are evenly spaced, with all ratios or spacings being different.

\paragraph{Data.}

We just need to specify the order and the dimension (e.g., $k=3$ and $n=8$).

\paragraph{Model.}

A first \p3 model, in a file `PerfectRythmingTiling1.py', used for the \x3 competition is:

\begin{boxpy}\begin{python}
@\imp@

k, n = data
m = k * n

max_gap = (m - 2) // (k - 1)

# x[j] is the value at the jth position of the sequence
x = VarArray(size=m, dom=range(n))

# gap[i] is the gap between successive occurrences of i
gap = VarArray(size=n, dom=range(1, max_gap + 1))

satisfy(
   # ensuring each value occurs k times
   Cardinality(x, occurrences={i: k for i in range(n)}),

   # all gaps are different
   AllDifferent(gap)
)

if not variant():
   
   # fst[i] is the first index where i occurs
   fst = VarArray(size=n, dom=range(m))

   satisfy(
      # all subsequences start at a different position (and are ordered)
      Increasing(fst, strict=True),  

      # ensuring subsequences follow an arithmetic progression
      [
         [fst[i] + gap[i] * (k - 1) < m for i in range(n)],
         [x[fst[i] + gap[i] * q] == i for i in range(n) for q in range(k)]
      ]
   )

elif variant('table'):

   # y[i][q] is the position of the qth occurrence of i
   y = VarArray(size=[n, k], dom=range(m))

   satisfy(
      # ensuring tiling
      AllDifferent(y),

      # all subsequences start at a different position (and are ordered)
      Increasing(y[:, 0], strict=True)
   )
    
   def table(i):
      return [(*pos, g, *[i if v in pos else ANY for v in range(m)]) for f in range(m)
                 for g in range(1, max_gap + 1) if f + g * (k - 1) < m
                     and (pos := [f + g * q for q in range(k)])]

   satisfy(
      (y[i], gap[i], x) in table(i) for i in range(n)
    )
\end{python}\end{boxpy}

This model involves three arrays of variables, and global constraints of type \gb{allDifferent}, \gb{cardinality}, and \gb{table}.
A series of $2 * 6 = 12$ instances has been selected for the competition (here, we have a main variant and a variant `table'). 
For generating an \x3 instance (file), you can execute for example:
\begin{command}
  python PerfectRythmingTiling1.py -data=[number,number]
  python PerfectRythmingTiling1.py -data=[number,number] -variant=table
\end{command}

A second \p3 model, in a file `PerfectRythmingTiling2.py', used for the \x3 competition is:

\begin{boxpy}\begin{python}
@\imp@

k, n = data
m = k * n

max_gap = (m - 2) // (k - 1)

# x[i][q] is the position of the qth occurrence of i
x = VarArray(size=[n, k], dom=range(m))

# gap[i] is the gap between successive occurrences of i
gap = VarArray(size=n, dom=range(1, max_gap + 1))

satisfy(
   # ensuring tiling
   AllDifferent(x),

   # all gaps are different
   AllDifferent(gap),

   # all subsequences start at a different position (and are ordered)
   Increasing(x[:, 0], strict=True)
)

if not variant():

   satisfy(
      # ensuring the same gap
      x[i][q + 1] - x[i][q] == gap[i] for i in range(n) for q in range(k - 1)
    )

elif variant('table'):

   T = [(*pos, g) for f in range(m) for g in range(1, max_gap + 1)
          if f + g * (k - 1) < m and (pos := [f + g * q for q in range(k)])]

   satisfy(
      # ensuring the same gap
      (x[i], gap[i]) in T for i in range(n)
   )
\end{python}\end{boxpy}

This model involves two arrays of variables, and global constraints of type \gb{allDifferent}, and \gb{table}.
A series of $2 * 6 = 12$ instances has been selected for the competition (here, we have a main variant and a variant `table'). 
For generating an \x3 instance (file), you can execute for example:
\begin{command}
  python PerfectRythmingTiling2.py -data=[number,number]
  python PerfectRythmingTiling2.py -data=[number,number] -variant=table
\end{command}

\subsection{RectPacking}

\paragraph{Description.}

The rectangle (square) packing problem consists of squares (boxes)
to be put in an enclosing rectangle (container) without overlapping of the squares.

\paragraph{Data.}

As an illustration of data specifying an instance of this problem, we have:
\begin{json}
{
  "container": {"width": 112, "height": 112},
  "boxes":[
    {"width": 2, "height": 2},
    {"width": 4, "height": 4},
    ..,
    {"width": 50, "height": 50}
  ]
} 
\end{json}

\paragraph{Model.}

The \p3 model, in a file `RectPacking.py', used for the \x3 competition is:

\begin{boxpy}\begin{python}
@\imp@

(width, height), boxes = data 

nBoxes = len(boxes)
B = range(nBoxes)

# x[i] is the x-coordinate where is put the ith rectangle
x = VarArray(size=nBoxes, dom=range(width))

# y[i] is the y-coordinate where is put the ith rectangle
y = VarArray(size=nBoxes, dom=range(height))

satisfy(
   # unary constraints on x
   [x[i] + boxes[i].width <= width for i in B],

   # unary constraints on y
   [y[i] + boxes[i].height <= height for i in B],

   # no overlap on boxes
   NoOverlap(
      origins=[(x[i], y[i]) for i in B],
      lengths=[(w, h) for (w, h) in boxes]
   ),

   # tag(symmetry-breaking)
   [
      x[-1] <= (width - boxes[-1].width) // 2,
      y[-1] <= x[-1]
   ] if width == height else None
)
\end{python}\end{boxpy}

This model involves two arrays of variables, and global constraints of type \gb{noOverlap}.
A series of 14 instances has been selected for the competition.
For generating an \x3 instance (file), you can execute for example:
\begin{command}
  python RectPacking.py -data=<datafile.json>
\end{command}

\subsection{Steiner3}

This is Problem \href{http://www.csplib.org/Problems/prob044}{044} on CSPLib.

\paragraph{Description.}

The ternary Steiner problem of order $n$ consists of finding a set of $n \times (n-1)/6$ triples of distinct integer 
elements in $\{1, 2, \dots,n \}$ such that any two triples have at most one common element.

\paragraph{Data.}

We just need to specify the order $n$ (e.g., $n=8$).

\paragraph{Model.}

The \p3 model, in a file `Steiner3.py', used for the \x3 competition is:

\begin{boxpy}\begin{python}
@\imp@

n = data or 6

nTriples = (n * (n - 1)) // 6

T = {(i1,i2,i3,j1,j2,j3) for (i1,i2,i3,j1,j2,j3) in product(range(1, n + 1), repeat=6)
      if different_values(i1, i2, i3) and different_values(j1, j2, j3)
         and len({i for i in {i1, i2, i3} if i in {j1, j2, j3}}) <= 1}

# x[i] is the ith triple of values
x = VarArray(size=[nTriples, 3], dom=range(1, n + 1))

satisfy(
   # each triple must be formed of strictly increasing integers
   [Increasing(triple, strict=True) for triple in x],

   # each pair of triples must share at most one value
   [(triple1 + triple2) in T for (triple1, triple2) in combinations(x, 2)]
)
\end{python}\end{boxpy}

This model involves a single array of variables, and global constraints of type \gb{table}.
A series of 8 instances has been selected for the competition.
For generating an \x3 instance (file), you can execute for example:
\begin{command}
  python Steiner3.py -data=<number>
\end{command}

\section{COP}

\subsection{BACPj}

\paragraph{Description.}

The goal is to assign periods to courses in a way that the academic load of each period is balanced, i.e., as similar as possible.
Here, we have a model with \gb{binPacking} constraints.

\paragraph{Data.}

As an illustration of data specifying an instance of this problem, we have:
\begin{json}
{
  "nCourses": 50,
  "nPeriods": 10,
  "loadLB": 2,
  "loadUB": 100,
  "coursesLB": 2,
  "coursesUB": 10,
  "loads": [6, 3, ..., 6],
  "prerequisites": [
    [2, 0],
    [3, 0],
    ...,
    [49, 46]
  ]
}
\end{json}

\paragraph{Model.}

The \p3 model, in a file `BACPj.py', used for the \x3 competition is:

\begin{boxpy}\begin{python}
@\imp@
    
nCourses, nPeriods, minCredits, maxCredits, minCourses, maxCourses, credits, prerequisites, factor = data if len(data) == 9 else (*data,1)

if factor > 1:  # arbitrarily increasing the problem
   q = nCourses
   nCourses *=factor
   nPeriods *=factor
   t = [v for v in credits]
   for _ in range(factor-1):
      t.extend(credits)
   credits = t #credits * factor
   prerequisites = [(i + q * k, j + q * k) for k in range(factor) for i, j in prerequisites]


# x[i] is the period (schedule) for the ith course
x = VarArray(size=nCourses, dom=range(nPeriods))

# co[j] is the number of courses at the jth period
co = VarArray(size=nPeriods, dom=range(minCourses, maxCourses + 1))

# cr[j] is the number of credits at the jth period
cr = VarArray(size=nPeriods, dom=range(minCredits, maxCredits + 1))

satisfy(
   # counting the number of courses in each period
   BinPacking(x, sizes=1, loads=co),

   # counting the number of credits in each period
   BinPacking(x, sizes=credits, loads=cr),

   # handling prerequisites
   [x[i1] < x[i2] for (i1, i2) in prerequisites],

   # tag(redundant)
   [
      Sum(co) == nCourses,
     
      Sum(x) == co * range(nPeriods),
     
      Sum(cr) == sum(credits)
   ]
)

minimize(
   # minimizing the sum of the square of the credit loads
   Sum(cr[j] * cr[j] for j in range(nPeriods))
)
\end{python}\end{boxpy}

This model involves three arrays of variables, and global constraints of type \gb{binPacking} and \gb{sum}.
A series of 15 instances has been selected for the competition.
For generating an \x3 instance (file), you can execute for example:
\begin{command}
  python BACPj.py -data=01.json
  python BACPj.py -data=[01.json,factor=4] 
\end{command}

\subsection{Clique}

\paragraph{Description.}

The problem is to find a clique of maximum size in a directed graph, specified by its adjacency matrix.

\paragraph{Data.}

As an illustration of data specifying an instance of this problem, we have:
\begin{json}
{
   "n": 5,
   "d": 5,
   "edges": [[0, 1], [1, 2], [2, 3], [3, 0], [2, 4], [3, 4]]
}
\end{json}

\paragraph{Model.}

The \p3 model, in a file `Clique.py', used for the \x3 competition is:

\begin{boxpy}\begin{python}
@\imp@

n, _, edges = data

matrix = [[0] * n for _ in range(n)]
for i, j in edges:
   matrix[i][j] = matrix[j][i] = 1


# x[i] is 1 iff the ith node belongs to the clique
x = VarArray(size=n, dom={0, 1})

satisfy(
    # the clique includes at most one of any pair of non-neighbouring vertices
    Sum(x[i], x[j]) < 2
      for i, j in combinations(n, 2) if matrix[i][j] == matrix[j][i] == 0
)

maximize(
    # maximizing the size of the clique
    Sum(x)
)
\end{python}\end{boxpy}

This model involves a single array of variables, and global constraints of type \gb{sum}.
A series of 15 instances has been selected for the competition.
For generating an \x3 instance (file), you can execute for example:
\begin{command}
  python Clique.py -data=toy.json
\end{command}

\subsection{CrosswordDesign}

\paragraph{Description.}

Given a main dictionary containing ordinary words, and a second dictionary containing thematic words, the objective is to fill up a grid with words of both dictionaries.
Each word from the thematic word has a value (benefit) equal to its length.
The objective is to maximize the overall value. The problem is difficult because black cells are not imposed, i.e., can be put anywhere in the grid
(but no adjacency of black cells is authorized).
See \cite{BB_romanian}.

\paragraph{Data.}

Data are given by two numbers (order and maximum number of words on rows and columns), a dictionary file, and a file with thematic words.

\paragraph{Model.}

The \p3 model, in a file `CrosswordDesign.py', used for the \x3 competition is:

\begin{boxpy}\begin{python}
@\imp@

n, nMaxWords, dictMain, dictThem = data
# n is the order of the grid (number of rows and number of columns)
# nMaxWords is the maximum number of words on a same row or column

nMaxWords = (n + 1) // 2 if nMaxWords == -1 else nMaxWords  # computed when necessary
wordSizeLimit = -1  
wordSizeLimit = n if wordSizeLimit == -1 else wordSizeLimit  # computed when necessary

mainWords = [alphabet_positions(word) for line in open(dictMain)
                 if len(word := line.strip().lower()) <= wordSizeLimit]
themWords = [alphabet_positions(word) for line in open(dictThem)
                 if len(word := line.strip().lower()) <= wordSizeLimit]

oneLetterWords = [(d,) for d in range(26)]
twoLetterWords = [(d1, d2) for d1 in range(26) for d2 in range(26)]

allWords = sorted(oneLetterWords + twoLetterWords + mainWords + themWords)
nWords = len(allWords)

ptsWords = [len(word) if word in themWords else 0 for word in allWords]

N_LETTERS, BLACK_POINT = 26, 26

def starred_table(k):
   lastWord = k == nMaxWords - 1
   T = []
   if k != 0:
      T.append((-1, -1, 0, -1, *[ANY] * n))
   possible_positions = [0, 1] if k == 0 else list(range(2 * k, n))
   for p in possible_positions:
      for i, word in enumerate(allWords):
         bp = p + len(word)  # position of the black point, just after the word
         if bp > n or (lastWord and bp < n - 1):
             continue
         t = [0] * (4 + n)
         t[0] = p
         t[1] = i
         t[2] = ptsWords[i]
         t[3] = -1 if lastWord else bp + 1 if bp <= n - 2 else -1
         for j in range(4, n + 4):
            if j - 4 in (p - 1, bp):
               t[j] = BLACK_POINT
            elif p <= j - 4 < p + len(word):
               t[j] = word[j - 4 - p]
            else:
               t[j] = ANY
         T.append(tuple(t))
   return T


Ts = [starred_table(k) for k in range(nMaxWords)]  # precomputed starred tables

# r[i][k] is the (index of) kth word at row i; -1 means 'no word'
r = VarArray(size=[n, nMaxWords], dom=range(-1, nWords))

# c[j][k] is the (index of) kth word at col j; -1 means 'no word'
c = VarArray(size=[n, nMaxWords], dom=range(-1, nWords))

# pr[i][k] is the position (index of col) of the kth word at row i; -1 if none
pr = VarArray(size=[n, nMaxWords + 1],
               dom=lambda i, k: {0, 1} if k == 0 else {-1} if k == nMaxWords else range(-1, n))

# pc[j][k] is the position (index of row) of the kth word at col j; -1 if none
pc = VarArray(size=[n, nMaxWords + 1],
               dom=lambda j, k: {0, 1} if k == 0 else {-1} if k == nMaxWords else range(-1, n))

# br[i][k] is the benefit of the kth word at row i
br = VarArray(size=[n, nMaxWords], dom=range(n + 1))

# bc[j][k] is the benefit of the kth word at col j
bc = VarArray(size=[n, nMaxWords], dom=range(n + 1))

# x[i][j] is the (number of) letter at row i and col j; 26 stands for a black point
x = VarArray(size=[n, n], dom=range(N_LETTERS + 1))  # +1 for black points

satisfy(
   # putting words on rows
   [
      Table(
         scope=(pr[i][k], r[i][k], br[i][k], pr[i][k + 1], x[i]),
         supports=Ts[k]
      ) for i in range(n) for k in range(nMaxWords)
   ],

   # putting words on columns
   [
      Table(
         scope=(pc[j][k], c[j][k], bc[j][k], pc[j][k + 1], x[:, j]),
         supports=Ts[k]
      ) for j in range(n) for k in range(nMaxWords)
   ]
)

maximize(
   Sum(br + bc)
)
\end{python}\end{boxpy}

This model involves seven arrays of variables, and global constraints of type \gb{table} and \gb{sum}.
A series of 20 instances has been selected for the competition.
For generating an \x3 instance (file), you can execute for example:
\begin{command}
  python CrosswordDesign.py -data=[10,4,"CrossMain.txt","CrossThem17.txt"] 
\end{command}

\subsection{EternityApprox}

\paragraph{Description.}

On a board of size n×m, you have to put square tiles (pieces) that are described by four colors (one for each direction : top, right, bottom and left).
All adjacent tiles on the board must have matching colors along their common edge. All edges must have color `0’ on the border of the board.

Here, this is a model with some form of relaxation.

\paragraph{Data.}
As an illustration of data specifying an instance of this problem, we have:
\begin{json}
{ 
   "n": 6,
   "m": 6,
   "pieces": [
     [0, 0, 1, 1],
     [0, 0, 1, 3],
     ...
     [5, 8, 6, 8]
  ]
}
\end{json}

\paragraph{Model.}

The \p3 model, in a file `EternityApprox.py', used for the \x3 competition is:

\begin{boxpy}\begin{python}
@\imp@

n, m, pieces = data

nPieces = len(pieces)
max_value = max(max(piece) for piece in pieces)  # max possible value on pieces

BAD1 = nPieces
BAD2 = nPieces * nPieces

rotations = [
   [(piece[r%4], piece[(r+1)%4], piece[(r+2)%4], piece[(r+3)%4]) for r in range(4)]
      for i, piece in enumerate(pieces)
]

if not variant():
   T = (
     [(i, complement(piece), ANY, ANY, ANY, BAD2) for i, piece in enumerate(pieces)]
     + [(i, ANY, complement(piece), ANY, ANY, BAD2) for i, piece in enumerate(pieces)]
     + [(i, ANY, ANY, complement(piece), ANY, BAD2) for i, piece in enumerate(pieces)]
     + [(i, ANY, ANY, ANY, complement(piece), BAD2) for i, piece in enumerate(pieces)]
     + [(i, *t, BAD1 if sorted(t) != sorted(piece) else 0 if t in rotations[i] else 1)
          for i, piece in enumerate(pieces) for t in product(piece, repeat=4)]
  )
elif variant("simp"):
   T = (
     [(i, *rotation, 0) for i, piece in enumerate(pieces) for rotation in rotations[i]]
     + [(i, complement(piece), ANY, ANY, ANY, BAD1) for i, piece in enumerate(pieces)]
     + [(i, ANY, complement(piece), ANY, ANY, BAD1) for i, piece in enumerate(pieces)]
     + [(i, ANY, ANY, complement(piece), ANY, BAD1) for i, piece in enumerate(pieces)]
     + [(i, ANY, ANY, ANY, complement(piece), BAD1) for i, piece in enumerate(pieces)]
     + [(i, inside(piece), inside(piece), inside(piece), inside(piece), 1)
          for i, piece in enumerate(pieces)]
   )

# x[i][j] is the index of the piece at row i and column j
x = VarArray(size=[n, m], dom=range(n * m))

# top[i][j] is the value at the top of the piece at row i and column j
top = VarArray(size=[n + 1, m], dom=range(max_value + 1))

# lft[i][j] is the value at the left of the piece at row i and column j
lft = VarArray(size=[n, m + 1], dom=range(max_value + 1))

z = VarArray(size=[n, m], dom={0, 1, BAD1, BAD2} if not variant() else {0, 1, BAD1})

satisfy(
   # all pieces must be placed (only once)
   AllDifferent(x),

   # all pieces must be valid
   [
      (x[i][j], top[i][j], lft[i][j + 1], top[i + 1][j], lft[i][j], z[i][j]) in T
      for i in range(n) for j in range(m)
   ],

   # putting special value 0 on borders
   [
      top[0] == 0,
      top[-1] == 0,  # meaning that we have 0 at the bottom border of the grid
      lft[:, 0] == 0,
      lft[:, -1] == 0  # meaning that we have 0 at the right border of the grid
   ],

   # tag(redundant)
   Cardinality(
      within=top[1:-1] + lft[:, 1:-1],
      occurrences={v: len([1 for piece in pieces for w in piece if w == v]) // 2 for v in range(1, max_value + 1)}
   )
)

minimize(
   Sum(z)
)
\end{python}\end{boxpy}

This model involves three arrays of variables, and global constraints of type \gb{allDifferent}, \gb{cardinality}, \gb{sum} and \gb{table}.
A series of $2 \times 8 = 16$ instances has been selected for the competition (here, we have a main variant, and a variant `simp').
For generating an \x3 instance (file), you can execute for example:
\begin{command}
  python EternityApprox.py -data=06x06.json
  python EternityApprox.py -data=06x06.json -variant=simp
\end{command}

\subsection{ExaminationTimetabling}

\paragraph{Description.}

Examination timetabling (ETT) is a classical problem that every university has to deal with on a regular basis.
Here, this is a formulation of ETT, which applies to Italian universities \cite{ett1,ett2}.
It involves many specific constraints and objectives.
For example, some exams are composed of separate written and oral part, which must be scheduled at suitable distance and have different overlap acceptability levels in relation to other exams.
In addition, the same exam might be repeated more than once in a session, with prescribed minimal distances among rounds.
As another quite distinctive feature, exams might require multiple rooms, typically in exclusive use.

\paragraph{Data.}
As an illustration of data specifying an instance of this problem, we have:
\begin{json}
{
  "events": {
    "roomedEvents": [1, 0, 1, 0, 1, 0, 1, 1, 1, 1, 1, 1, 1, 1, 0],
    "conflictLevels": [
      [0, -1, -1, -1, -1, -1, -1, -1, 0, 0, -1, -1, -1, -1, -1],
      [-1, 0, -1, -1, -1, -1, -1, -1, 0, 0, -1, -1, -1, -1, -1],
      ...
      [-1, -1, -1, -1, -1, -1, 5, 5, 0, 0, 5, 5, 5, 5, 0]
    ],
    "distanceWeights": [
      [0, 15, 12, 0, 12, 0, 2, 2, 0, 0, 2, 0, 2, 0, 2],
      [15, 0, 0, 0, 0, 0, 0, 0, 0, 0, 0, 0, 0, 0, 0],
      ...
      [2, 0, 2, 0, 2, 0, 2, 2, 0, 0, 2, 0, 2, 0, 0]
    ],
    "minDistances": [
      [0, 4, 5, 5, 5, 5, 6, 6, 0, 0, 6, 6, 6, 6, 2],
      [0, 0, 5, 5, 5, 5, 6, 6, 0, 0, 6, 6, 6, 6, 2],
      ...
      [2, 2, 2, 2, 2, 2, 2, 2, 0, 0, 2, 2, 2, 2, 0]
    ],
    "maxDistances": [
      [20, 8, 20, 20, 20, 20, 20, 20, 20, 20, 20, 20, 20, 20, 20],
      [20, 20, 20, 20, 20, 20, 20, 20, 20, 20, 20, 20, 20, 20, 20],
      ...
      [20, 20, 20, 20, 20, 20, 20, 20, 20, 20, 20, 20, 20, 20, 20]
    ],
    "precedences": [
      [0, 1, 1, 1, 1, 1, 0, 0, 0, 0, 0, 0, 0, 0, 0],
      [0, 0, 1, 1, 1, 1, 0, 0, 0, 0, 0, 0, 0, 0, 0],
      ...
      [0, 0, 0, 0, 0, 0, 0, 0, 0, 0, 0, 0, 0, 0, 0]
    ],
    "eventPeriodConstraints": [
      [-1, 0, 0, 0, -1, 0, 0, 0, 0, -1, 0, 0, 0, 0, 0, 0, 0, 0, 0, 0],
      [-1, 0, 0, 0, -1, 0, 0, 0, 0, 10, 0, 0, 0, 0, 0, 0, 0, 0, 0, 0],
      ...
      [0, 0, 0, 0, -1, 0, 0, 0, 0, 10, 0, 0, 0, 0, 0, 0, 0, 0, 0, 0]
    ],
    "eventRoomConstraints": [
      [-1, -1, -1, -1, -1, 0, 0, -1],
      [-1, -1, -1, -1, -1, -1, -1, 0],
      ...
      [-1, -1, -1, -1, -1, -1, -1, 0]
    ]
  },
  "rooms": {
    "roomPeriodConstraints": [
      [0, 0, 0, 0, 0, 0, 0, 0, 0, 0, 0, 0, 0, 0, 0, 0, 0, 0, 0, -1],
      [0, 0, 0, 0, 0, 0, 0, 0, 0, 0, 0, 0, 0, 0, 0, 0, 0, 0, 0, 0],
      ...
      [0, 0, 0, 0, 0, 0, 0, 0, 0, 0, 0, 0, 0, 0, 0, 0, 0, 0, 0, 0]
    ],
    "roomsetOverlaps": [
      [1, 0, 0, 0, 0, 1, 1, 0],
      [0, 1, 0, 0, 0, 1, 0, 0],
      ...
      [0, 0, 0, 0, 0, 0, 0, 0]
    ]
  }
}
\end{json}

\paragraph{Model.}

A first \p3 model, in a file `ExaminationTimetablig1.py', used for the \x3 competition is:

\begin{boxpy}\begin{python}
@\imp@

events, rooms = data 

roomedEvents, conflictLevels, distanceWeights, minDistances, maxDistances, precedences, epConstraints, erConstraints = events
rpConstraints, roomsetOverlaps = rooms

nEvents, nRooms, nPeriods = len(roomedEvents), len(rpConstraints), len(rpConstraints[0])
E, R, P, HARD = range(nEvents), range(nRooms), range(nPeriods), -1

present_dummy_room = any(roomedEvents[e] == 0 for e in E)
DUMMY_ROOM = nRooms - 1 if present_dummy_room else -1

EC1, EC2 = equivalent_event_classes(events)
# code for equivalent_event_classes not given here

# ep[i] is the period for the ith event
ep = VarArray(size=nEvents, dom=range(nPeriods))

# er[i] is the room for the ith event
er = VarArray(size=nEvents, dom=range(nRooms))


C = [(e1, e2) for e1, e2 in combinations(E, 2) if conflictLevels[e1][e2] != HARD
      and not precedences[e1][e2] and not precedences[e2][e1]
      and roomedEvents[e1] and roomedEvents[e2]]

if len(C) > 0:
   R1, R2 = (R[:-1], R[:-1]) if present_dummy_room else (R, R)  # :-1 for the dummy room 
   real_overlap = any((roomsetOverlaps[r1][r2] == 1) != (r1==r2) for r1 in R1 for r2 in R2)
   if not real_overlap:
      satisfy(
         either(
            ep[e1] != ep[e2],
            er[e1] != er[e2]
         ) for e1, e2 in C
      )
   else:
      satisfy(
         either(
            ep[e1] != ep[e2],
            roomsetOverlaps[er[e1]][er[e2]] == 0
         ) for e1, e2 in C
      )


satisfy(
   # respecting hard conflicts
   [ep[e1] != ep[e2] for e1, e2 in combinations(E, 2) if conflictLevels[e1][e2] == HARD],

   # respecting precedences
   [ep[e1] < ep[e2] for e1, e2 in combinations(E, 2) if precedences[e1][e2]],

   # ensuring rooms are available at chosen periods
   [
      Table(
         scope=(er[e], ep[e]),
         conflicts=T
      ) for e in E
   ] if len(T := [(r,p) for r in R for p in P if rpConstraints[r][p] == HARD])>0 else None,

   # forbidding certain periods for certain events
   [ep[e] not in [p for p in P if epConstraints[e][p] == HARD] for e in E],

   # forbidding certain rooms for certain events
   [
      er[e] not in [r for r in R if erConstraints[e][r] == HARD]
      if roomedEvents[e] else er[e] == DUMMY_ROOM
      for e in E
   ],

   # tag(symetry-breaking)
   [Increasing(ep[cl]) for cl in EC1]
)

er3Constraints = cp_array([max(0, v) for v in row] for row in erConstraints)
ep3Constraints = cp_array([max(0, v) for v in row] for row in epConstraints)
rp3Constraints = cp_array([max(0, v) for v in row] for row in rpConstraints)
    
minimize(
   Sum(
     (ep[e1] == ep[e2]) * conflictLevels[e1][e2]
       for e1, e2 in combinations(E, 2) if conflictLevels[e1][e2] > 0)  # ConflictCost
   + Sum(
     distanceWeights[e1][e2] * max(0, minDistances[e1][e2] - ep[e2] + ep[e1])
       for e1 in E for e2 in E if precedences[e1][e2]
         and distanceWeights[e1][e2] > 0 and minDistances[e1][e2] > 0
   )  # MinDirectionalDistanceCost
   + Sum(
      distanceWeights[e1][e2] * max(0, ep[e2] - ep[e1] - maxDistances[e1][e2])
        for e1 in E for e2 in E if precedences[e1][e2]
          and distanceWeights[e1][e2] > 0 and maxDistances[e1][e2] < nPeriods
   )  # MaxDirectionalDistanceCost
   + Sum(
      distanceWeights[e1][e2] * max(0, minDistances[e1][e2] - abs(ep[e2] - ep[e1]))
        for e1, e2 in combinations(E, 2) if not precedences[e1][e2]
          and distanceWeights[e1][e2] > 0 and minDistances[e1][e2] > 0
   )  # MinUndirectionalDistanceCost
   + Sum(
      distanceWeights[e1][e2] * max(0, abs(ep[e2] - ep[e1]) - maxDistances[e1][e2])
        for e1, e2 in combinations(E, 2) if not precedences[e1][e2]
          and distanceWeights[e1][e2] > 0 and maxDistances[e1][e2] < nPeriods
   )  # MaxUndirectionalDistanceCost
   + Sum(er3Constraints[e][er[e]] for e in E)  # RoomPreferenceCost
   + Sum(ep3Constraints[e][ep[e]] for e in E)  # PeriodPreferenceCost
   + Sum(rp3Constraints[er[e]][ep[e]] for e in E)  # RoomPeriodCost
)
\end{python}\end{boxpy}

This model involves two arrays of variables, and global constraints of type \gb{element}, \gb{sum} and \gb{table}.
A series of 15 instances has been selected for the competition.
For generating an \x3 instance (file), you can execute for example:
\begin{command}
  python ExaminationTimetabling1.py -data=<datafile.json> 
\end{command}

A second \p3 model, in a file `ExaminationTimetablig3.py', has been used for the \x3 competition.

\subsection{Feedback}

\paragraph{Description.}

Given a directed graph, the feedback arc set problem \cite{CGSSZW_soft} consists in removing a minimum subset of the arcs in order to obtain an acyclic subgraph.
An alternative formulation is to find a total order $<$ on the vertices such that there is a minimum number of feedback arcs (i.e. an arc from i to j with $i > j$).
In order to experiment with submodular problems, we modified the penalty function so that if there is a feedback arc (i,j), instead of having a cost of 1 we have a cost proportional to the difference between the ordering positions of i and j.
The resulting problem is similar to a simple temporal CSP with linear preferences.
    
\paragraph{Data.}

As an illustration of data specifying an instance of this problem, we have:
\begin{json}
{
  "n": 50,
  "arcs": [
    [0, 1],
    [0, 9],
    ...,
    [49, 40]
  ]
}
\end{json}

\paragraph{Model.}

The \p3 model, in a file `Feedback.py', used for the \x3 competition is:

\begin{boxpy}\begin{python}
@\imp@

n, arcs = data
e = len(arcs)

# x[i] is the position (order) of the ith node
x = VarArray(size=n, dom=range(n))

# z[k] is the penalty of the kth arc
z = VarArray(size=e, dom=range(n))

satisfy(
   # ensuring a total order on nodes
   AllDifferent(x),

   # computing penalties related to feedback arcs
   [z[k] == ift(x[i] < x[j], 0, x[i] - x[j]) for k, (i, j) in enumerate(arcs)]
)

minimize(
   # minimizing the sum of penalties
   Sum(z)
)
\end{python}\end{boxpy}

This model involves two arrays of variables, and global constraints of type \gb{allDifferent} and \gb{sum}.
A series of 20 instances has been selected for the competition.
For generating an \x3 instance (file), you can execute for example:
\begin{command}
  python Feedback.py -data=P0050-100.json
\end{command}

\subsection{Golomb Ruler}

This is Problem \href{http://www.csplib.org/Problems/prob006}{006} on CSPLib.

\paragraph{Description.}

\begin{quote}
The problem is to find the ruler with the smallest length where we can put $n$ marks such that the distance between any two pairs of marks is distinct.
\end{quote}

\paragraph{Data.}
As an illustration of data specifying an instance of this problem, we simply have $n=8$.

\paragraph{Model.}

The \p3 model, in a file `GolombRuler.py', used for the \x3 competition is:

\begin{boxpy}\begin{python}
@\imp@

assert variant("a3") or variant("a4")

n = data or 8

ub = 200 if n < 10 else (n // 10) * 500  # a rather large upper-bound of an optimal ruler

# x[i] is the position of the ith tick
x = VarArray(size=n, dom=range(ub+1))

if variant("a4"):

   satisfy(
      abs(x[j] - x[i]) != abs(x[q] - x[k])
         for i, j in combinations(range(n), 2) for k, q in combinations(range(i + 1, n), 2)
   )

elif variant("a3"):
         
   # y[i][j] is the distance between x[i] and x[j], for i strictly less than j
   y = VarArray(size=[n, n], dom=lambda i, j: range(1, ub+1) if i < j else None)

   satisfy(
      # tag(channeling)
      [x[j] == x[i] + y[i][j] for i, j in combinations(n, 2)],

      AllDifferent(y),
   )

satisfy(
   # tag(symmetry-breaking)
   [
      x[0] == 0,
      Increasing(x, strict=True)
   ]
)

minimize(
   x[-1]
)
\end{python}\end{boxpy}

This model involves one or two arrays of variables, according to the chosen variant (`a3' or `a4'), and global constraints of type \gb{allDifferent}.
A series of $2 \times 7 = 14$ instances has been selected for the competition (here, we have a variant `a3' and a variant `a4').
For generating an \x3 instance (file), you can execute for example:
\begin{command}
  python GolombRuler.py -data=10 -variant=a3
  python GolombRuler.py -data=10 -variant=a4 
\end{command}

\subsection{MaximumAgreementForest}

\paragraph{Description.}

This is the problem of the challenge in 2026 for PACE (Parameterized Algorithms and Computational Experiments, \href{https://pacechallenge.org/2026/}{https://pacechallenge.org/2026/}):
\begin{itemize}
\item as input, we have  a list of phylogenetic trees on the same leaf-set,
\item as output, we want an agreement forest for all input trees with a minimum number of trees.
\end{itemize}

\paragraph{Data.}

As an illustration of data specifying an instance of this problem, we have:
\begin{json}
{
  "nLeaves": 6,
  "trees": [
    [[[5, 6], [3, 4]], [1, 2]],
    [[[[[4, 2], 1], 5], 3], 6]
  ]
}
\end{json}

\paragraph{Model.}

The \p3 model, in a file `MaximumAgreementForest2.py', used for the \x3 competition is:

\begin{boxpy}\begin{python}
@\imp@

nLeaves, input_trees = data 

trees, all_arcs, arcs_leaving, arcs_leading, descendants, leaves_nodes, deps, reachable_leaves = build_structures(nLeaves, input_trees)
# code for build_structures not given here

nodes = DUMMY = [len(tree.parents) for tree in trees]  # DUMMY[t] is the last dummy value
L, L0, V = max(nodes), nodes[0], range(1, nLeaves + 1)  # ! values of leaves start at 1
N, T = [[i for i in range(nodes[t]) if not is_leaf(trees[t], i)] for t in range(len(input_trees))], range(len(input_trees))


def domain_mp(i, t):
   if t == 0:
      return None
   if is_leaf(trees[0], i):
      return {leaves_nodes[t][trees[0].children[i]]}
   pv = set(reachable_leaves[0][i])
   return [j for j in N[t] if len(pv.intersection(reachable_leaves[t][j])) >= 2] + [DUMMY[t]]


# nds[t][i] is 1 if the ith node of tree t is selected (dummy node at last position)
nds = VarArray(
   size=[T, L + 1],
   dom=lambda t, i: None if i > nodes[t] else {0} if i == DUMMY[t] else {1}
                      if is_leaf(trees[t], i) else {0, 1}
)

# acs[t][k] is 1 if the kth arc of tree t is selected
acs = VarArray(
   size=[T, max(len(t) for t in all_arcs)],
   dom=lambda t, k: None if k >= len(all_arcs[t]) else {0, 1}
)

# rts[i] is 1 if the ith node of tree t is a root
rts = VarArray(
   size=[T, L + 1],
   dom=lambda t, i: None if i > nodes[t] else {0} if i == DUMMY[t] else {0, 1}
)

# sld[i] is 1 if the ith node of tree t is solid
sld = VarArray(
   size=[T, L + 1],
   dom=lambda t, i: None if i > nodes[t] else {0} if i == DUMMY[t] else {1}
                      if is_leaf(trees[t], i) else {0, 1})  
)

# sds[t][i] is the first solid descendant node of the ith node (possibly itself)
# and itself if not selected
sds = VarArray(
   size=[T, L + 1],
   dom=lambda t, i: None if i > nodes[t] else {i} if i == DUMMY[t] else descendants[t][i]
)

# mp[i][t] is the node in tree t mapped from the ith node of tree 0
mp = VarArray(size=[L0, T], dom=domain_mp)

# ml[i][t] is the left son of the node in tree t mapped from the ith node of tree 0
ml = VarArray(size=[L0, T], dom=lambda i, t: None if t == 0 else range(nodes[t] + 1))

# mr[i][t] is the right son of the node in tree t mapped from the ith node of tree 0
mr = VarArray(size=[L0, T], dom=lambda i, t: None if t == 0 else range(nodes[t] + 1))

# sml[i][t] is the first solid descendant node of the left son of the node in tree t
# mapped from the ith node of tree 0
sml = VarArray(size=[L0, T], dom=lambda i, t: None if t == 0 else range(nodes[t] + 1))

# smr[i][t] is the first solid descendant node of the right son of the node in tree t
# mapped from the ith node of tree 0
smr = VarArray(size=[L0, T], dom=lambda i, t: None if t == 0 else range(nodes[t] + 1))

# z is the number of roots  (in each tree)
z = Var(dom=V)

satisfy(
   # ensuring coherence of arcs : (0,1,*) and (*,1,0) not permitted
   [
      (
         either(acs[t][k] == 0, nds[t][i] == 1),
         either(acs[t][k] == 0, nds[t][j] == 1)
      ) for t in T for k, (i, j) in enumerate(all_arcs[t])
   ],

   # at least one arc leaving a selected node
   [
      disjunction(nds[t][i] == 0, acs[t][k1] == 1, acs[t][k2] == 1)
      for t in T for i in N[t] if (k1 := arcs_leaving[t][i][0], k2 := arcs_leaving[t][i][1])
   ],

   # ensuring coherence of leaving arcs
   [
      [AllEqual(nds[t][0], acs[t][arcs_leaving[t][0][0]], acs[t][arcs_leaving[t][0][1]])
         for t in T],
      [
         If(
            acs[t][arcs_leading[t][i]] == 0,
            Then=AllEqual(nds[t][i], acs[t][k1], acs[t][k2])
         ) for t in T for i in N[t][1:]
             if (k1 := arcs_leaving[t][i][0], k2 := arcs_leaving[t][i][1])
      ]
   ],

   # computing roots
   [
      (
         rts[t][0] == nds[t][0],
         [rts[t][i] == both(nds[t][i] == 1, acs[t][arcs_leading[t][i]] == 0)
            for i in range(1, nodes[t])]  # N[t][1:]]
      ) for t in T
   ],

   # roots are solid nodes
   [
      If(
         rts[t][i],
         Then=sld[t][i]  # [acs[t][k1] == 1, acs[t][k2] == 1]
      ) for t in T for i in N[t]
   ],

   # computing solid descendants
   [
      Table(
         scope=(nds[t][i], acs[t][k1], acs[t][k2], sld[t][i],sds[t][i],sds[t][j1],sds[t][j2]),
         supports=[
            (0, 0, 0, 0, i, j1, j2),
            (1, 1, 1, 1, i, ANY, ANY),
            ((1, 1, 0, 0, v, v, j2) for v in descendants[t][j1]),
            ((1, 0, 1, 0, v, j1, v) for v in descendants[t][j2])
         ]
      ) for t in T for i in N[t]
           if (k1 := arcs_leaving[t][i][0], k2 := arcs_leaving[t][i][1])
               and (j1 := trees[t].children[i][0], j2 := trees[t].children[i][1])
   ],

   # mapping to different nodes
   [AllDifferent([mp[i][t] for i in N[0]], excepting=DUMMY[t]) for t in T[1:]],

   # mapping only solid nodes to not dummy nodes  tag(redundant)
   [sld[0][i] == (mp[i][t] != DUMMY[t]) for i in N[0] for t in T[1:]],

   # mapping solid nodes to solid nodes
   [sld[t][mp[i][t]] == sld[0][i] for i in N[0] for t in T[1:]],

   # mapping roots to roots
   [rts[t][mp[i][t]] == rts[0][i] for i in N[0] for t in T[1:]],

   # computing left and right children of mapped nodes
   [
      Table(
         scope=(ml[i][t], mr[i][t], mp[i][t]),
         supports=[
            ((j1, j2, j) for j in mp[i][t].dom if j != DUMMY[t]
               and (j1 := trees[t].children[j][0], j2 := trees[t].children[j][1])),
            (DUMMY[t], DUMMY[t], DUMMY[t])
         ]
      ) for i in N[0] for t in T[1:]
   ],

   # computing real mapped children
   [
      (
         sml[i][t] == sds[t][ml[i][t]],
         smr[i][t] == sds[t][mr[i][t]],
      ) for i in N[0] for t in T[1:]
   ],

   # ensuring coherent mapping
   [
      disjunction(
         sld[0][i] == 0,
         both(mp[sds[0][j1]][t] == sml[i][t], mp[sds[0][j2]][t] == smr[i][t]),
         both(mp[sds[0][j1]][t] == smr[i][t], mp[sds[0][j2]][t] == sml[i][t])
      ) for i in N[0] for t in T[1:]
           if (j1 := trees[0].children[i][0], j2 := trees[0].children[i][1])
   ],

   # ensuring the same number of roots in each input tree
   [z == Sum(rts[t]) for t in T],

   # tag(redundant)
   [Sum(acs[t]) + z + rts[t][0] <= len(all_arcs[t]) for t in T],

   # tag(r1)
   [AllEqual(rts[t][leaves_nodes[t][v]] for t in T) for v in V],

   # tag(r2)
   [
      If(
         rts[t][i],
         Then=[rts[tt][j] == 1 for (tt, j) in lst]
      ) for (t, i), lst in deps.items()
   ]
)

minimize(
   # minimizing the number of roots
   z
)
\end{python}\end{boxpy}

This model involves ten arrays of variables, and global constraints of type \gb{allDifferent}, \gb{allEqual}, \gb{element}, \gb{table} and \gb{sum}.
A series of 20 instances has been selected for the competition.
For generating an \x3 instance (file), you can execute for example:
\begin{command}
  python MaximumAgreementForest2.py -data=<datafile.json> 
\end{command}

\subsection{MicroplateSwapping}

\paragraph{Description.}

From \cite{GRSS_plate}:
\begin{quote}
  Given a list of experiments to be tested, the problem consists of assigning every experiment to a destination plate to perform the following procedure:
  \begin{enumerate}
\item an empty destination plate is placed in the dispensing robot,
\item an appropriate source plate is placed in the dispensing robot,
\item  the robot dispenses compounds from wells of the source plate to the assigned wells of the destination plate,
\item  if the destination plate has all its wells filled with the planned experiments (i.e. it is serviced), then:
  \begin{itemize}
\item if all destination plates are serviced, terminate the process,
\item  else, return to step 1 to swap to the next empty destination plate.
  \end{itemize}
  Otherwise, return to step 2 (swap the source plate).
  \end{enumerate}
\end{quote}

\paragraph{Data.}

As an illustration of data specifying an instance of this problem, we have:
\begin{json}
{
  "nTargetPlates": 2,
  "sizeTargetPlates": 4,
  "combinations": [
    [0, 1], [0, 1], [0, 1], [0, 1], [0, 1], [1, 1], [0, 1], [1, 1]
  ]
}
\end{json}

\paragraph{Model.}

The \p3 model, in a file `MicroplateSwapping.py', used for the \x3 competition is:

\begin{boxpy}\begin{python}
@\imp@

nTargetPlates, sizeTargetPlates, combinations = data 

nCombinations, nSourcePlates = len(combinations), len(combinations[0])
C, S, T = range(nCombinations), range(nSourcePlates), range(nTargetPlates)

used_source_plates = [i for i in S if any(comb[i] for comb in combinations)]

# tc[j][k] is 1 if the kth combination is used by the jth target plate
tc = VarArray(size=[nTargetPlates, nCombinations], dom={0, 1})

# st[i][j] is 1 if the ith source plate is connected to the jth target plate
st = VarArray(size=[nSourcePlates, nTargetPlates], dom={0, 1})

# nst[j] is the number of source plates to which is connected the jth target plate
nst = VarArray(size=nTargetPlates, dom=range(nSourcePlates + 1))

# cst[j] is the source plate connection steps
cst = VarArray(size=nTargetPlates - 1, dom=range(nSourcePlates + 1))

satisfy(
   # each combination must be used by one target plate
   [Sum(tc[:, k]) == 1 for k in C],

   # the number of assigned combinations to a single target plate must not exceed its size
   [
      If(
         sizeTargetPlates * nTargetPlates == nCombinations,
         Then=[Sum(tc[j]) == sizeTargetPlates for j in T]
      ),
      If(
         sizeTargetPlates * nTargetPlates > nCombinations,
         Then=[Sum(tc[j]) <= sizeTargetPlates for j in T]
      )
   ],

   # determining whether source plates and target plates are connected
   [st[i][j] == Exist(tc[j][k] for k in C if combinations[k][i])
      for i in used_source_plates for j in T],

   # computing the number of connections of target plates
   [nst[j] == Sum(st[:, j]) for j in T],

   [
      If(
         cst[j] >= 1,
         Then=[
            st[cst[j]][j],
            st[cst[j]][j + 1]
         ]
      ) for j in T[:-1]
   ],

   [
      If(
         AllHold(st[i][j] * st[i][j + 1] == 0 for i in S),
         Then=cst[j] == 0
      ) for j in T[:-1]
   ],

   [
      If(
         cst[j] >= 1, nst[j + 1] > 1,
         Then=cst[j + 1] != cst[j]
      ) for j in T[:-2]
   ]
)

minimize(
   Sum(st) - Sum(cst[j] >= 1 for j in T[:-1])
)

\end{python}\end{boxpy}

This model involves four arrays of variables, and global constraints of type \gb{element} and \gb{sum}.
A series of 20 instances has been selected for the competition.
For generating an \x3 instance (file), you can execute for example:
\begin{command}
python MicroplateSwapping.py -data=<datafile.json>
python MicroplateSwapping.py -data=<datafile.dzn> -parser=MicroplateSwapping_ParserZ.py
\end{command}

\subsection{NonDominatingQueens}

\paragraph{Description.}

The problem is to place queens on an $n$ by $n$ board so that the total number of squares attacked/occupied by the queens is minimized.

\paragraph{Data.}

We just need to specify the order of the boad (e.g., $n=8$).

\paragraph{Model.}

The \p3 model, in a file `NonDominatingQueens.py', used for the \x3 competition is:

\begin{boxpy}\begin{python}
@\imp@

n = data 

ATTACKED, FREE = 0, 1

symmetries = [f.apply_on(n) for f in TypeSquareSymmetry]

def attacked_by(i, j, binary=False):
   if binary:
      return [ATTACKED if (k == i or q == j or abs(k - i) == abs(q - j))
                         else ANY for k in range(n) for q in range(n)]
   return [k * n + q for k in range(n) for q in range(n)
              if k == i or q == j or abs(k - i) == abs(q - j)]


# x[k] is the cell position of the kth queen
x = VarArray(size=n, dom=range(n * n))

# y[i][j] is 1 if the cell at row i and column j is not attacked by a queen
y = VarArray(size=[n, n], dom={ATTACKED, FREE})

satisfy(
   # putting queens on different cells (and in increasing order)
   Increasing(x, strict=True),

   # tag(symmetry-breaking)
   [y <= y[symmetry] for symmetry in symmetries]
)

if not variant():
   
   satisfy(
      # handling attacked cells
      [
         If(
            Exist(x[k].among(attacked_by(i, j)) for k in range(n)),
            Then=y[i][j] == ATTACKED
         ) for i in range(n) for j in range(n)
      ]
   )

elif variant("table"):

   T = [(i * n + j, *attacked_by(i, j, True)) for i in range(n) for j in range(n)]

   satisfy(
      # handling attacked cells
      [(x[k], y) in T for k in range(n)]
   )

maximize(
   # maximizing the number of free cells
   Sum(y)
)
\end{python}\end{boxpy}

This model involves two arrays of variables, and global constraints \gb{count}, \gb{lex}, \gb{table} and \gb{sum}.
A series of $2 \times 12 = 24$ instances has been selected for the competition (here, we have a main variant and a variant `table').
For generating an \x3 instance (file), you can execute for example:
\begin{command}
  python NonDominatingQueens.py -data=<number> 
  python NonDominatingQueens.py -data=<number> -variant=table
\end{command}

\subsection{Satellite Observations}

\paragraph{Description.}

From \cite{VL_satellite} and \href{https://forge.inrae.fr/thomas.schiex/cost-function-library/-/tree/master/crafted/bep?ref_type=heads}{cost-function-library}:
\begin{quote}
This is a simplified version of a problem of selecting and scheduling observations for agile satellites.

The satellite has a pool of candidate photographs to take, and must select and schedule a subset of them, at each pass above a strip of the earth territory.
The satellite can only take one photograph at a time.
A photograph can only be taken  during a given time window, depending on its coordinates on the earth surface.
Minimal time manoeuvres are required between two consecutive photographs.
All physical constraints (time windows, time manoeuvres) must be met, and the sum of  the revenues of the selected photographs must be maximized.

This problem is a mix between a TSP with time windows, and a Knapsack problem.
The main components of an instance are :
\begin{enumerate}
\item a pool of photo candidates, each with a duration, a time window and a revenue,
\item a 2D array of manoeuvre times beetween each pair of photo candidate.
\end{enumerate}

Original version by François Laburthe as a benchmark for CHOCO (oct 2000).
Adaptation for an EOLE bench by Michel Lemaître (june 2001).
\end{quote}
    
\paragraph{Data.}

As an illustration of data specifying an instance of this problem, we have:
\begin{json}
{
  "photos": [
    [54, 294, 494, 210],
    [40, 97, 297, 204],
    [44, 454, 654, 176],
    [59, 413, 613, 234],
    [34, 618, 818, 172],
    [34, 143, 343, 211],
    [34, 301, 501, 71],
    [45, 118, 318, 218],
    [49, 196, 396, 155],
    [39, 873, 1073, 205]
  ],
  "times": [
    [0, 49, 42, 32, 81, 43, 17, 44, 25, 145],
    [49, 0, 90, 80, 130, 25, 54, 5, 25, 194],
    [42, 90, 0, 10, 41, 78, 38, 85, 65, 104],
    [32, 80, 10, 0, 51, 68, 28, 75, 54, 115],
    [81, 130, 41, 51, 0, 119, 80, 125, 105, 63],
    [43, 25, 78, 68, 119, 0, 39, 25, 20, 182],
    [17, 54, 38, 28, 80, 39, 0, 50, 28, 143],
    [44, 5, 85, 75, 125, 25, 50, 0, 21, 189],
    [25, 25, 65, 54, 105, 20, 28, 21, 0, 169],
    [145, 194, 104, 115, 63, 182, 143, 189, 169, 0]
  ]
}
\end{json}

\paragraph{Model.}

The \p3 model, in a file `SatelliteObservations.py', used for the \x3 competition is:

\begin{boxpy}\begin{python}
@\imp@

hotos, times = data or None

durations, earliest, latest, revenues = zip(*photos)

nPhotos, nSteps = len(photos), len(photos)
T = range(nSteps)

UNDEF_TIME = 1_000_000  # because in data we may have some negative earliest time

# x[i] is the photo taken at step i, or -1 if none
x = VarArray(size=nSteps, dom=range(-1, nPhotos))

# dur[i] is the duration to take the photo at step i, or 0 if none
dur = VarArray(size=nSteps, dom={0, durations})

# erl[i] is the earliest time to take the photo at step i
erl = VarArray(size=nSteps, dom={UNDEF_TIME, earliest})

# lst[i] is the latest time to take the photo at step i
lst = VarArray(size=nSteps, dom={UNDEF_TIME, latest})

# rev[i] is the revenue for taking the photo at step i, or 0 if none
rev = VarArray(size=nSteps, dom={0, revenues})

# beg[i] is the starting (beginning) time to take the photo at step i
beg = VarArray(size=nSteps, dom={UNDEF_TIME, range(min(earliest), max(latest) + 1)})

# mnr[i] is maneuver time between the two photos at step i and i+1
mnr = VarArray(size=nSteps - 1, dom=[0, times])

Ta = [(-1,ANY),(ANY,-1)] + [(u,v) for u in range(nPhotos) for v in range(nPhotos) if u !=v]

Tb = [(-1, 0, UNDEF_TIME, UNDEF_TIME, 0)] + [tuple(flatten(t)) for t in enumerate(photos)]

satisfy(
   # taking different photos
   AllDifferent(x, excepting=-1),

   # computing information related to the photos taken
   [
      Table(
         scope=(x[i], dur[i], erl[i], lst[i], rev[i]),
         supports=Tb
      ) for i in T
   ],

   # taking photos in their allowed time intervals
   [
      (
         erl[i] <= beg[i],
         beg[i] + dur[i] <= lst[i]
      ) for i in T
   ],

   # computing manoeuvre times
   [
      Table(
         scope=(x[i], x[i + 1], mnr[i]),
         supports=[(ANY, -1, 0)] + [(i, j, times[i][j]) for i in range(nSteps) for j in range(nSteps) if i != j]
      ) for i in T[:-1]
   ],

   # ensuring photos can be possibly taken in sequence
   [beg[i] + dur[i] + mnr[i] <= beg[i + 1] for i in T[:-1]]
)

maximize(
   # maximizing global revenue
   Sum(rev)
)
\end{python}\end{boxpy}

This model involves seven arrays of variables, and global constraints of type \gb{allDifferent}, \gb{sum} and \gb {table}.
A series of 16 instances has been selected for the competition.
For generating an \x3 instance (file), you can execute for example:
\begin{command}
  python SatelliteObservations.py -data=<datafile.json>
  python SatelliteObservations.py -data=<datafile.txt> -parser=SatelliteObservations_Parser.py
\end{command}

\subsection{Sonet2}

\paragraph{Description.}

From \cite{S_symmetry,NAGJMS_automatically}:
\begin{quote}
  The SONET problem is a network design problem where each node is connected to a set of rings (ﬁbre-optic connections).
  The simpliﬁed SONET problem has the following parameters: the number of nodes n, the upper
  limit on the number of rings m, the maximum number of nodes per ring r, and a set of pairs that must be connected.
  For each of these pairs there must exist a ring connected to both nodes.
  The number of node-ring connections is minimised.
\end{quote}
    
\paragraph{Data.}

As an illustration of data specifying an instance of this problem, we have:
\begin{json}
{
  "nRings": 10,
  "nNodes": 6,
  "r": 3,
  "demand": [
    [0, 1, 1, 1, 0, 0],
    [0, 0, 0, 0, 0, 0],
    [0, 0, 0, 1, 0, 1],
    [0, 0, 0, 0, 0, 0],
    [0, 0, 0, 0, 0, 1],
    [0, 0, 0, 0, 0, 0]
  ]
}
\end{json}

\paragraph{Model.}

The \p3 model, in a file `Sonet2.py', used for the \x3 competition is:

\begin{boxpy}\begin{python}
@\imp@

Rings, nNodes, r, demands = data 

assert all(demands[i][j] == 0 for i in range(nNodes) for j in range(i + 1))

# x[i][k] is 1 if the ith ring contains the kth node
x = VarArray(size=[nRings, nNodes], dom={0, 1})

satisfy(
   # if there is a demand between 2 nodes, then there has to exist  a ring,
   # on which they are both installed
   [
      Exist(
         x[i][j1] + x[i][j2] >= 2 for i in range(nRings)
      ) for j1, j2 in combinations(nNodes, 2)  
   ],

   # respecting the capacity of rings
   [Sum(x[i]) <= r for i in range(nRings)],

   # tag(symmetry-breaking)
   LexIncreasing(x)
)

minimize(
   # minimizing the number of nodes installed on rings
   Sum(x)
)
\end{python}\end{boxpy}

This model involves a single array of variables, and global constraints of type \gb{count}, \gb{lex} and \gb{sum}.
A series of 16 instances has been selected for the competition.
For generating an \x3 instance (file), you can execute for example:
\begin{command}
  python Sonet2.py -data=<datafile.json>
  python Sonet2.py -data=<datafile.txt> -parser=Sonet2_ParserE.py
\end{command}

\subsection{SumSquares}

\paragraph{Description.}

From \href{https://www.janestreet.com/puzzles/sum-of-squares-index/}{JaneStreet}:
\begin{quote}
You have to place a digit in each of the spots of a $n \times n$ grid, so that each $n$-digit number (leading zeroes are ok) reading across and reading down is divisible by the number associated with it, trying to maximize the sum of the $n*n$ numbers you enter.
\end{quote}

\paragraph{Data.}

We just need to specify the order of the boad (e.g., $n=8$).

\paragraph{Model.}

The \p3 model, in a file `SumSquares.py', used for the \x3 competition is:

\begin{boxpy}\begin{python}
@\imp@

n = data 

# x[i][j] is the value at row i and column j
x = VarArray(size=[n, n], dom=range(10))

p = list(reversed([10 ** j for j in range(n)]))

satisfy(
   # each number reading across is divisible by the number associated with the row
   [(x[i] * p) % (i + 1) == 0 for i in range(1, n)],
  
   # each number reading down is divisible by the number associated with the column
   [(x[:, i - n] * p) % (i + 1) == 0 for i in range(n, 2 * n)]
)

maximize(
   # minimizing the sum of used numbers
   Sum(x)
)
\end{python}\end{boxpy}

This model involves a single array of variables, and global constraints of type \gb{sum}.
A series of 8 instances has been selected for the competition.
For generating an \x3 instance (file), you can execute for example:
\begin{command}
  python SumSquares.py -data=<number>
\end{command}

\subsection{TelecomScheduling}

\paragraph{Description.}

France Telecom Workforce Scheduling Problem submitted to \href{https://roadef.org/challenge/2007/en/index.php}{ROADEF 2007}.
\begin{quote}
It consists of scheduling a large set of maintenance interventions on the telecom network (from routine tasks to urgent repairs) over a multi-week horizon,
using a limited workforce of technicians who have individual skills, domains of expertise, and work calendars:
\begin{itemize}
\item each intervention requires specific competencies and possibly several technicians working in parallel,
\item some tasks must be sequenced after others,
\item and a team must stay together for the whole day.
\end{itemize}
The goal is to build feasible daily schedules for every technician that satisfy all these constraints  while minimizing the total priority-weighted completion time of the interventions.
\end{quote}

\paragraph{Data.}

As an illustration of data specifying an instance of this problem, we have:
\begin{json}
{
  "nDomains": 3,
  "nLevels": 2,
  "interventions": [
    {"duration": 60, "predecessors": [], "priority": 1, "cost": 400,
      "requirements": [[0, 0], [0, 0], [1, 0]]},
    {"duration": 60, "predecessors": [], "priority": 3, "cost": 400,
      "requirements": [[0, 0], [0, 0], [1, 0]]},
    {"duration": 15, "predecessors": [], "priority": 1, "cost": 100,
      "requirements": [[1, 0], [1, 0], [1, 1]]},
    {"duration": 15, "predecessors": [], "priority": 2, "cost": 300,
      "requirements": [[0, 0], [0, 0], [1, 1]]},
    {"duration": 60, "predecessors": [], "priority": 1, "cost": 150,
      "requirements": [[0, 0], [2, 0], [0, 0]]}
  ],
  "technicians": [
    {"skills": [1, 2, 2], "absences": [1]},
    {"skills": [1, 2, 2], "absences": [0, 2]},
    {"skills": [2, 2, 1], "absences": [3]},
    {"skills": [1, 2, 1], "absences": [4]},
    {"skills": [1, 1, 2], "absences": []}
  ],
  "budget": 0
}
\end{json}

\paragraph{Model.}

The \p3 model, in a file `TelecomScheduling.py', used for the \x3 competition is:

\begin{boxpy}\begin{python}
@\imp@

nSkills, nSkillLevels, tasks, agents, budget = data
nTasks, nAgents = len(tasks), len(agents)
durations, predecessors, priorities, costs, requirements = zip(*tasks)
assert all(1 <= v for v in priorities)  

ABSENT, DAYTIME_LIMIT, NB_PRIORITIES = 0, 120, max(priorities)

nTeams, nDays = nAgents, nTasks  # one should be able to compute best values (bounds)
T, A, D = range(nTasks), range(nAgents), range(nDays)

Ps = [[k for k, task in enumerate(tasks) if task.priority == p + 1]
         for p in range(NB_PRIORITIES)]  # tasks per priority (three subsets)

similar_tasks = [(k, q) for k, q in combinations(nTasks, 2) if (i1:=tasks[k], i2:=tasks[q])
                    and i1.duration == i2.duration and i1.predecessors == i2.predecessors
                    and i1.requirements == i2.requirements]
similar_tasks = [(k, q) if tasks[k].priority <= tasks[q].priority else (q, k)
                    for k, q in similar_tasks]

similar_agents = [(k, q) for k, q in combinations(nAgents, 2)
                    if agents[k].skills == agents[q].skills]


def domain_ts():
   S1 = set(durations)
   S2 = [v for v in S1 if not any(v % w == 0 for w in S1 if w != v)]
   S3 = sorted({v * i for v in S2 for i in range(DAYTIME_LIMIT + 1) if v*i <= DAYTIME_LIMIT})
   r_limit = number_of_values_for_sum_ge(S3, DAYTIME_LIMIT)
   S4 = sorted({sum(t) for r in range(r_limit + 1) for t in combinations(S3, r)
              if sum(t) <= DAYTIME_LIMIT})
   return S4


# x[i][d] is the team (index) of the ith agent on the dth day
x = VarArray(size=[nAgents, nDays], dom=nTeams)

# td[k] is the day chosen for the kth task
td = VarArray(size=nTasks, dom=nDays)

# ts[k] is the starting time for the kth task
ts = VarArray(size=nTasks, dom=domain_ts())  

# tt[k] is the team chosen for the kth task
tt = VarArray(size=nTasks, dom=nTeams)

# z[i] is the ending time of the ith subset of tasks
z = VarArray(size=4, dom=range(nDays * DAYTIME_LIMIT + 1))

satisfy(
   # not working agents are absent (in team 0)
   [x[i][d] == ABSENT for i in A for d in agents[i].absences if d < nDays],

   # tasks must be finished during the day
   [ts[k] + durations[k] <= DAYTIME_LIMIT for k in T],

   # handling absent agents (team 0)
   [(td[k], ts[k], tt[k]) in [(0, 0, ABSENT), (ANY, ANY, ne(ABSENT))] for k in T],

   # respecting the budget for non planned tasks
   Sum((tt[k] == ABSENT) * costs[k] for k in T) <= budget,

   # no overlapping tasks (same team same day)
   [
      If(
         tt[k] != ABSENT, tt[k] == tt[q], td[k] == td[q],
         Then=either(ts[k] + durations[k] <= ts[q], ts[q] + durations[q] <= ts[k])  
      ) for k, q in combinations(nTasks, 2)
   ],

   # respecting precedences
   [
      disjunction(
         tt[q] == ABSENT,
         tt[k] == ABSENT,
         td[q] < td[k],
         both(td[q] == td[k], ts[q] + durations[q] <= ts[k])
      ) for k in T for q in predecessors[k]
   ],

   # respecting coherence of not planned tasks
   [
      If(
         tt[q] == ABSENT,
         Then=tt[k] == ABSENT
      ) for k in T for q in predecessors[k]
   ],

   # ensuring enough agents with required skills
   [
      If(
         tt[k] != ABSENT,
         Then=Count(tt[k] == x[i][td[k]] for i in [i for i, agent in enumerate(agents)
             if agent.skills[sk] > sl]) >= requirements[k][sk][sl]
      ) for k in range(nTasks) for sk in range(nSkills) for sl in range(nSkillLevels)
         if requirements[k][sk][sl] > 0 and (sl == nSkillLevels - 1
             or requirements[k][sk][sl] != requirements[k][sk][sl + 1])
   ],

   # computing ending times
   [
      z[0] == Maximum(z[1], z[2], z[3]),
      [z[i + 1] == (0 if len(P) == 0 else Maximum(td[k] * DAYTIME_LIMIT + ts[k] + durations[k] for k in P)) for i, P in enumerate(Ps)]
   ],

   # tag(symmetry-breaking)
   [Precedence(x[:, d], values=[v for v in range(nTeams) if v != ABSENT]) for d in D],

   # every agent is declared absent after the day of the last task
   [
      If(
         d > z[0] // DAYTIME_LIMIT,
         Then=[x[i][d] == ABSENT for i in A]
      ) for d in D
   ],

   # tag(symmetry-breaking,similar)
   [
      x[i][d] <= x[j][d] for i, j in similar_agents for d in D if d not in agents[i].absences and d not in agents[j].absences
   ]
)

minimize(
   z * [1, 28, 14, 4]
)
\end{python}\end{boxpy}

This model involves five arrays of variables, and global constraints of type \gb{count}, \gb{maximium}, \gb{precedence},  \gb{sum}, and \gb{table}.
A series of 14 instances has been selected for the competition.
For generating an \x3 instance (file), you can execute for example:
\begin{command}
  python TelecomScheduling.py -data=<datafile.json>
  python TelecomScheduling.py -data=<directory> -parser=TelecomScheduling_Parse.py
\end{command}

\chapter{Solvers}

In this chapter, we introduce the solvers and teams having participated to the \x3 Competition 2026.

\begin{itemize}
\item ACE (Christophe Lecoutre)
\item Choco (Charles Prud'homme)
\item CoSoCo (Gilles Audemard)
\item CPMpy\_* {\small (Tias Guns, Thomas Sergeys, Ignace Bleukx, Hendrik Bierlee)} \\
  cpmpy\_ortools, cpmpy\_cpo, cpmpy\_z3, cpmpy\_gurobi, \\
  cpmpy\_highs, cpmpy\_scip,
  cpmpy\_exact, cpmpy\_oindakaas\_cadical
\item kat (Takehide Soh)
\item MaxiCP (Pierre Schaus, Guillaume Derval, Augustin Delecluse, Laurent Michel, and Pascal Van Hentenryck)
\item Nacre (Ga\"el Glorian)
\item Picat (Neng-Fa Zhou)
\item PyCSP3-solvers-extra (Sohaib Afifi)
\item Qayd (Sohaib Afifi)
\item Sat4j-CSP-PB (extension of Sat4j by Thibault Falque and Romain Wallon)
\item Tempo (Valentin Antuori, Emmanuel Hebrard, Tim Luchterhand) 
\item toulbar2 (David Allouche, Simon de Givry  et al.) 
\end{itemize}

\addcontentsline{toc}{section}{\numberline{}ACE}
\includepdf[pages=-,pagecommand={\thispagestyle{plain}}]{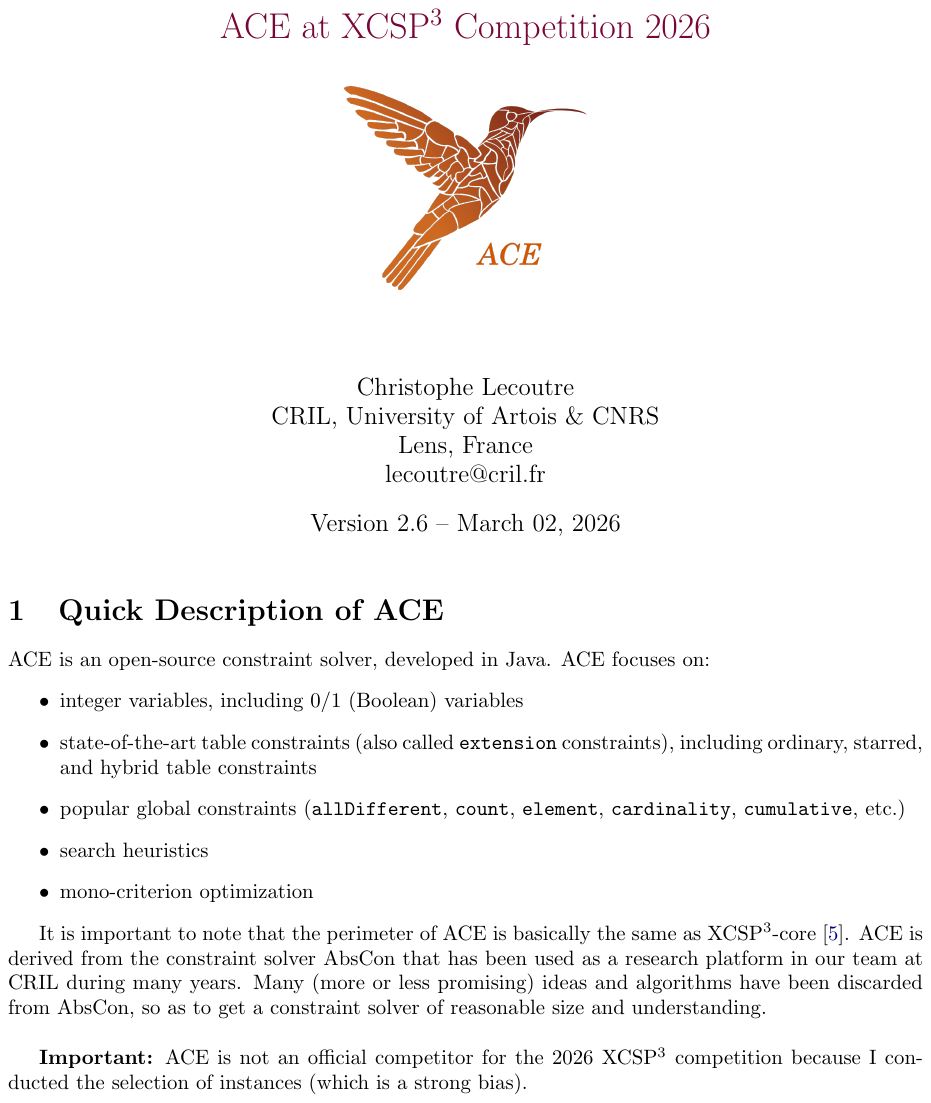}
\addcontentsline{toc}{section}{\numberline{}Choco}
\includepdf[pages=-,pagecommand={\thispagestyle{plain}}]{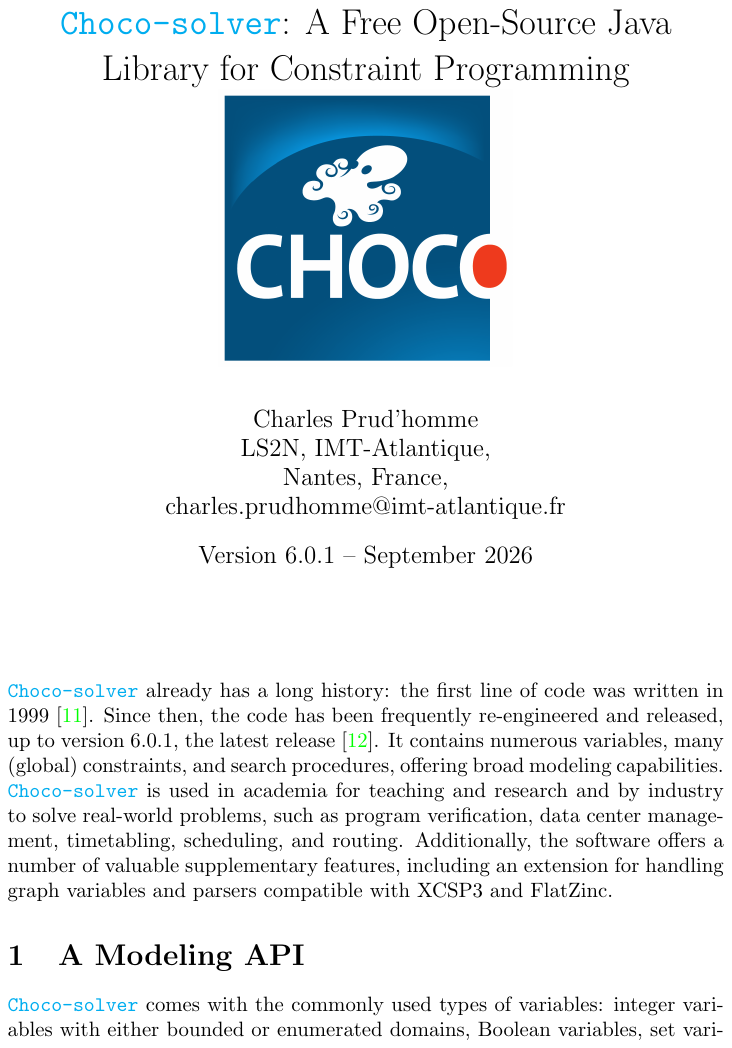}
\addcontentsline{toc}{section}{\numberline{}CoSoCo}
\includepdf[pages=-,pagecommand={\thispagestyle{plain}}]{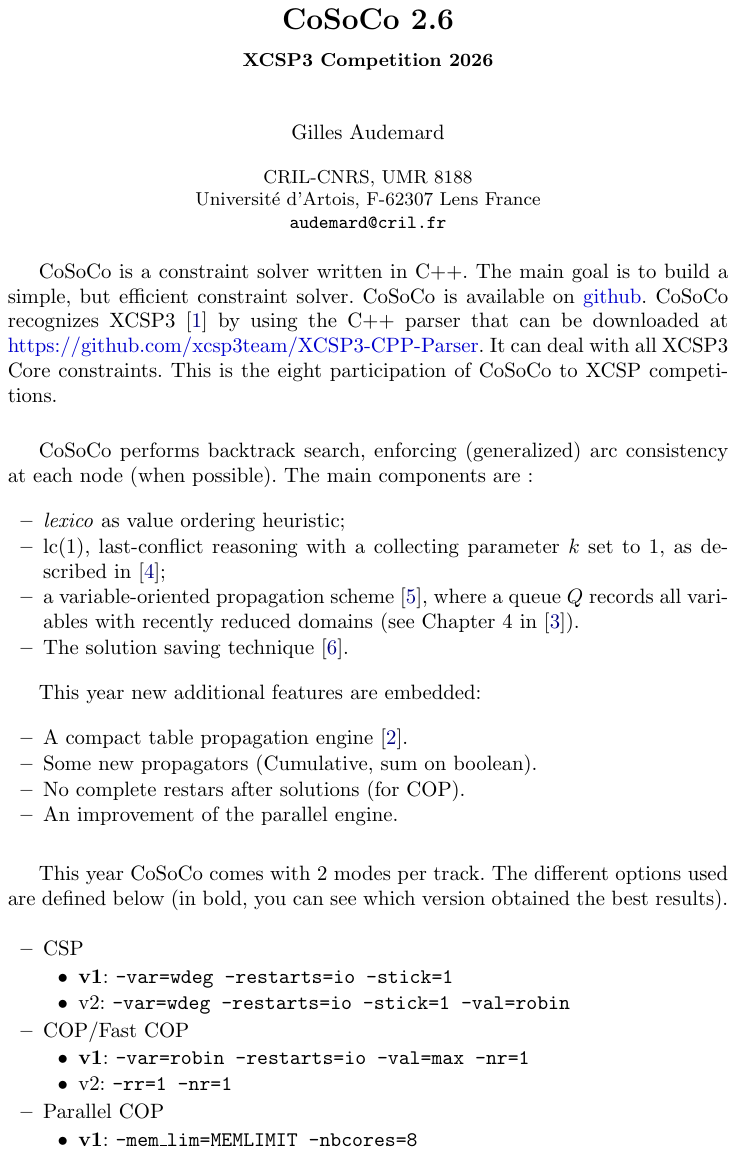}
\addcontentsline{toc}{section}{\numberline{}CPMpy-*}
\includepdf[pages=-,pagecommand={\thispagestyle{plain}}]{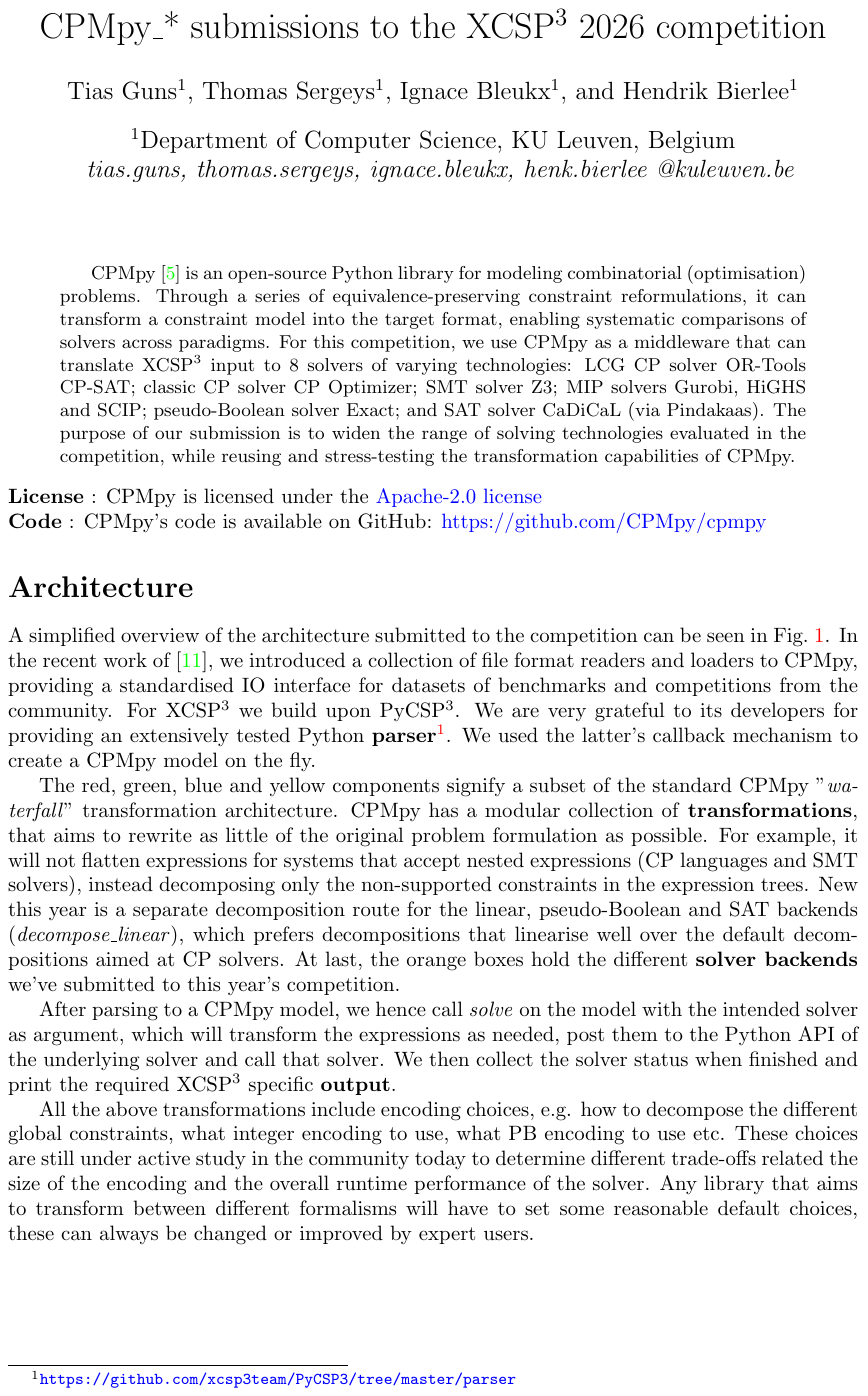}
\addcontentsline{toc}{section}{\numberline{}kat}
\includepdf[pages=-,pagecommand={\thispagestyle{plain}}]{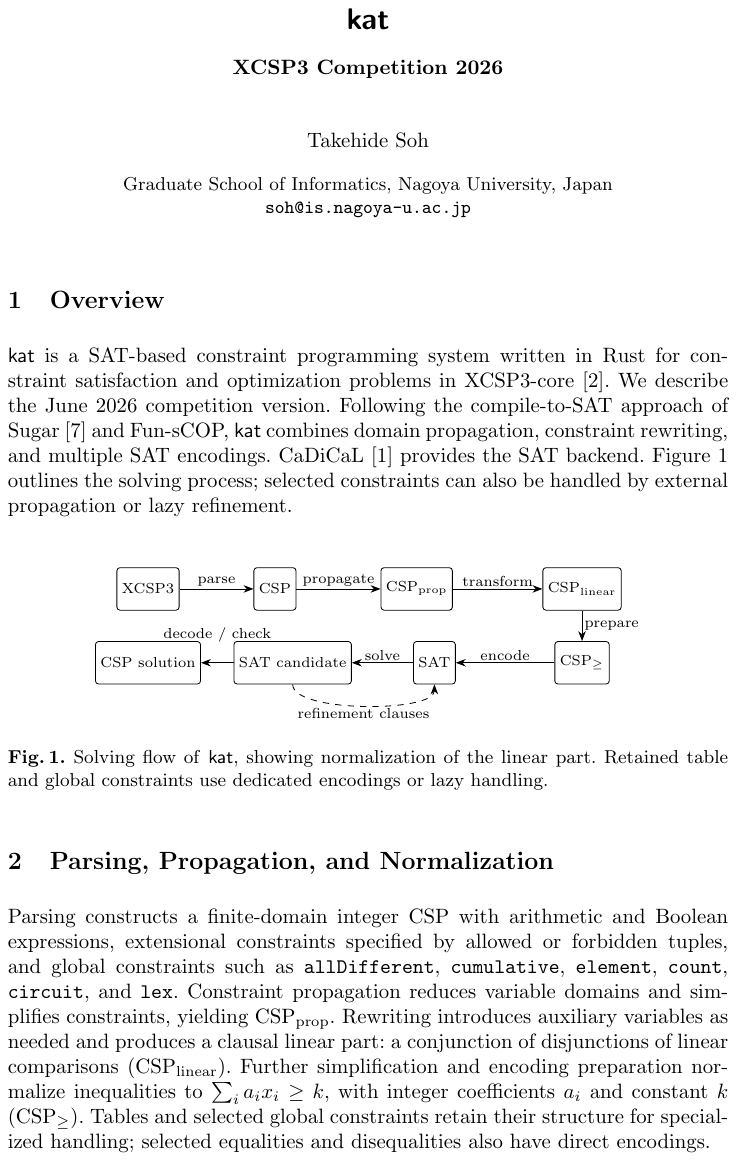}
\addcontentsline{toc}{section}{\numberline{}MaxiCP}
\includepdf[pages=-,pagecommand={\thispagestyle{plain}}]{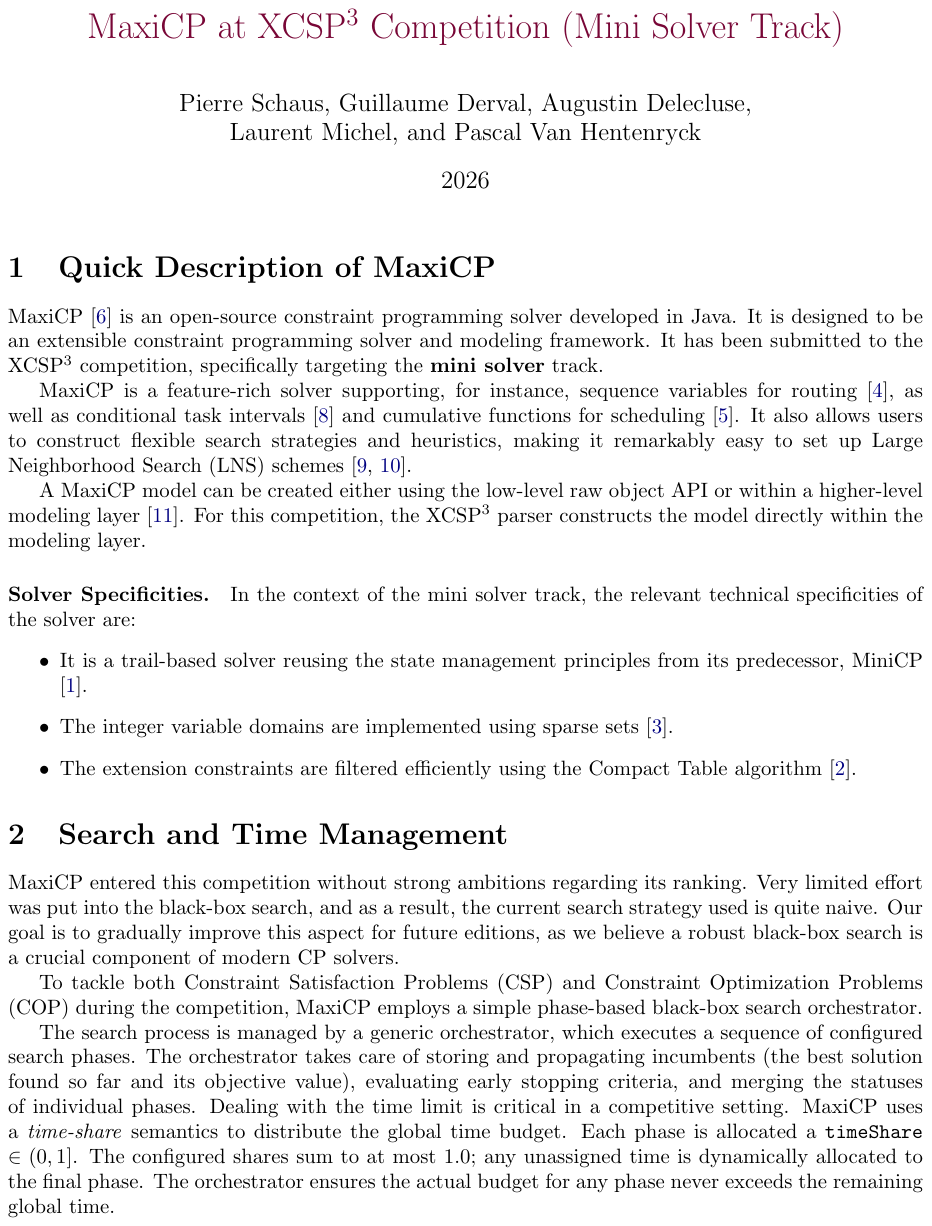}
\addcontentsline{toc}{section}{\numberline{}Nacre}
\includepdf[pages=-,pagecommand={\thispagestyle{plain}}]{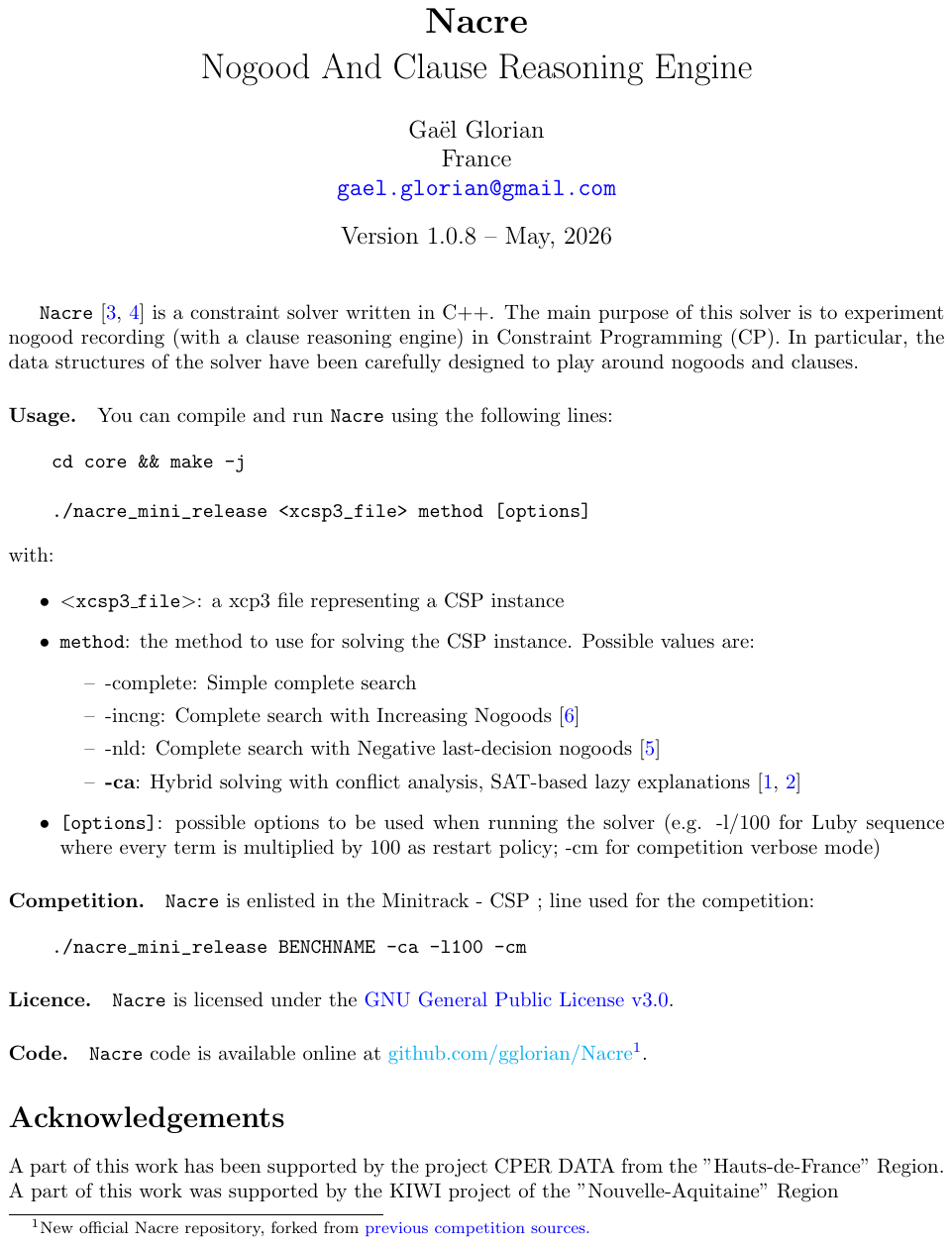}
\addcontentsline{toc}{section}{\numberline{}Picat}
\includepdf[pages=-,pagecommand={\thispagestyle{plain}}]{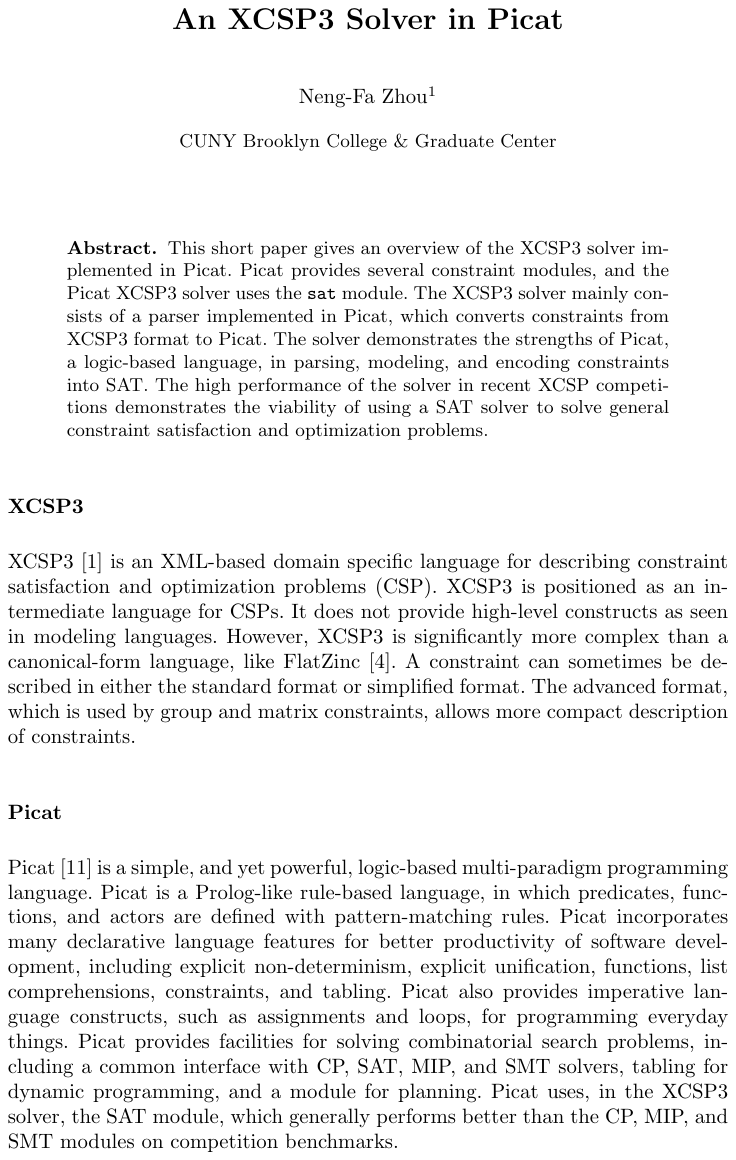}
\addcontentsline{toc}{section}{\numberline{}PyCSP3-solvers-extra}
\includepdf[pages=-,pagecommand={\thispagestyle{plain}}]{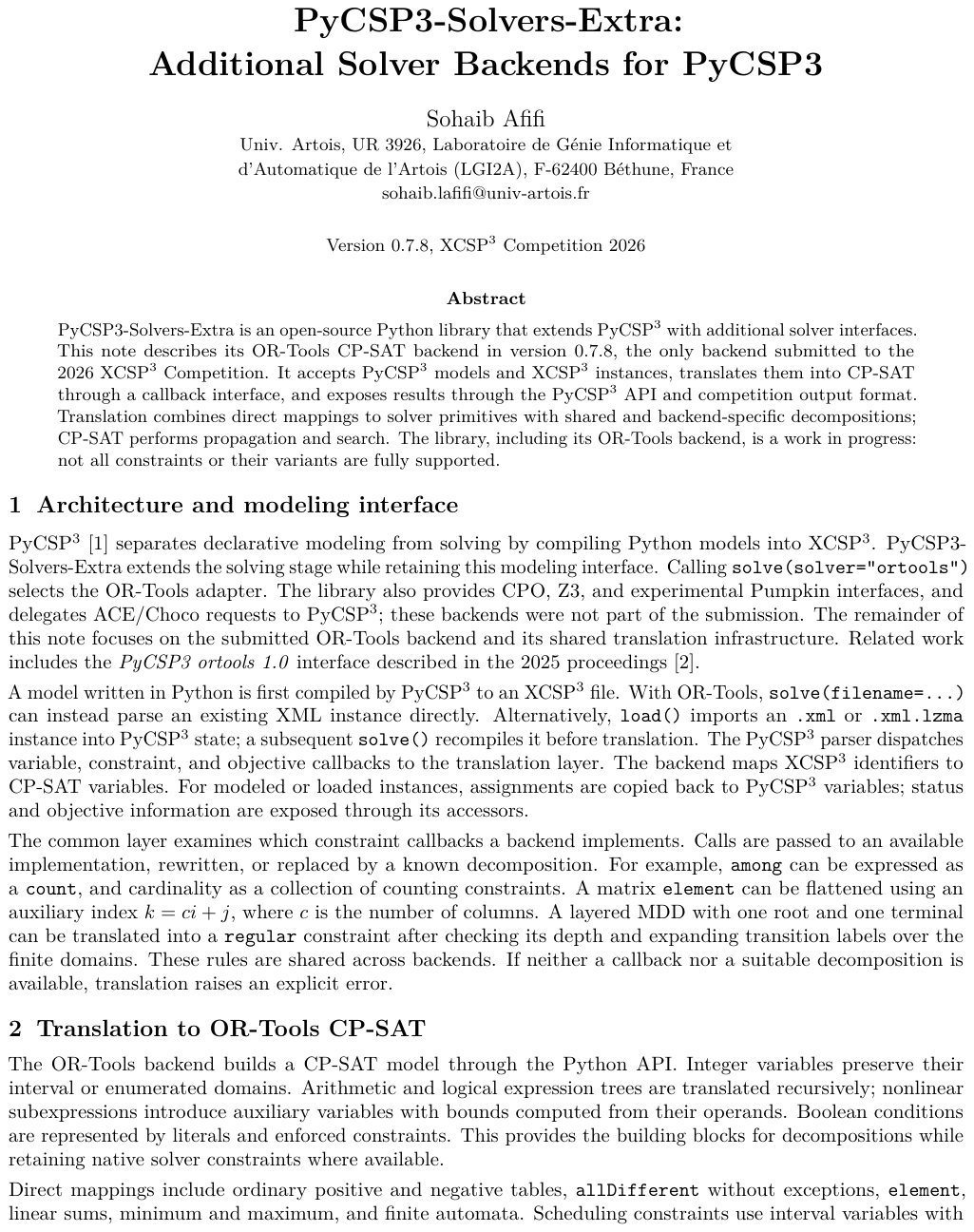}
\addcontentsline{toc}{section}{\numberline{}Qayd}
\includepdf[pages=-,pagecommand={\thispagestyle{plain}}]{Qayd.pdf}
\addcontentsline{toc}{section}{\numberline{}Sat4j-CSP-PB}
\includepdf[pages=-,pagecommand={\thispagestyle{plain}}]{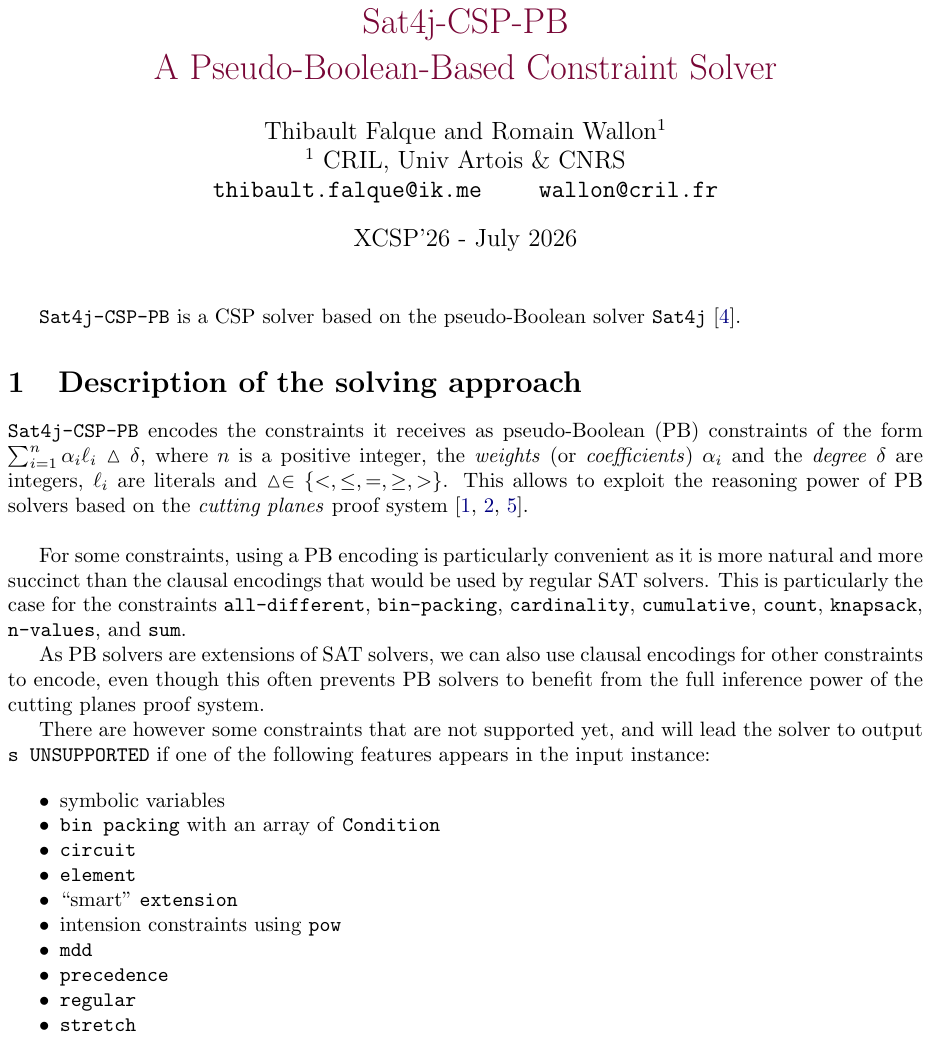}
\addcontentsline{toc}{section}{\numberline{}Tempo}
\includepdf[pages=-,pagecommand={\thispagestyle{plain}}]{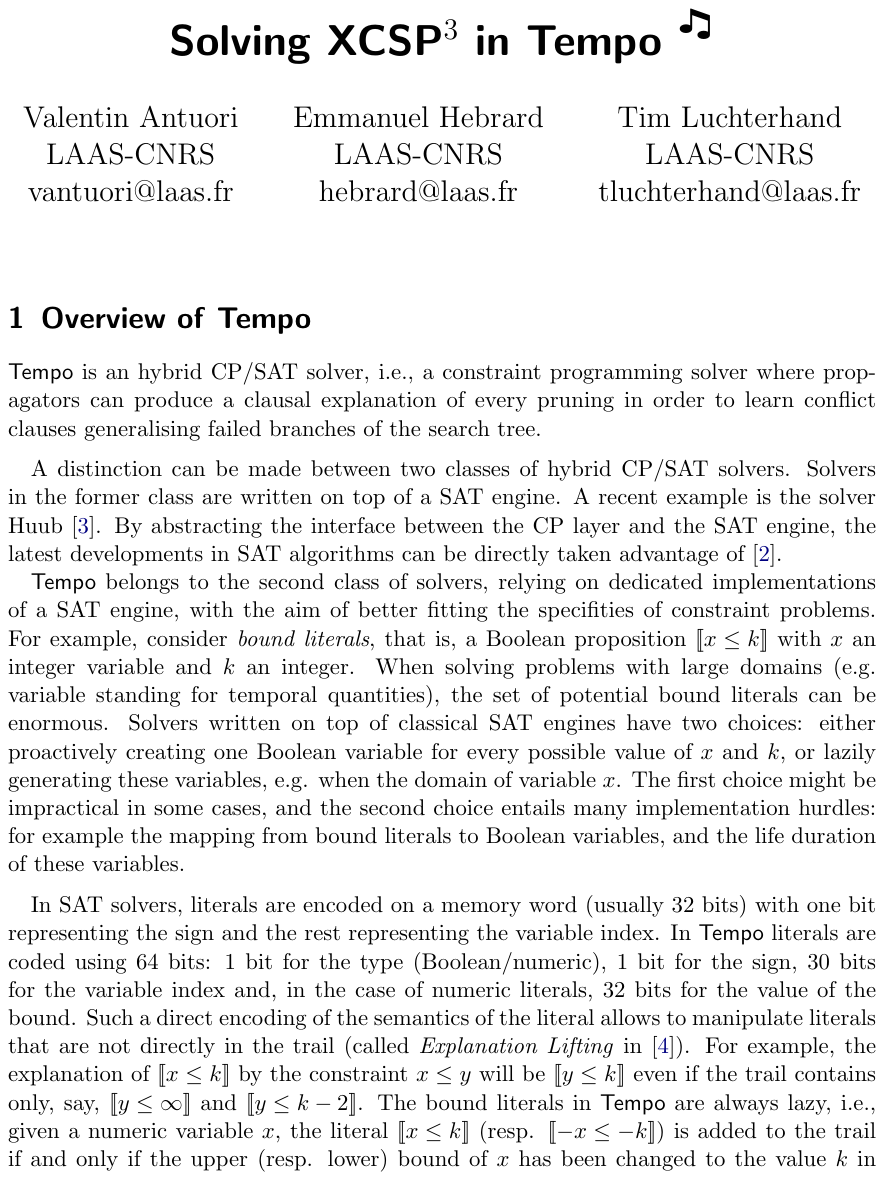}
\addcontentsline{toc}{section}{\numberline{}toulbar2}
\includepdf[pages=-,pagecommand={\thispagestyle{plain}}]{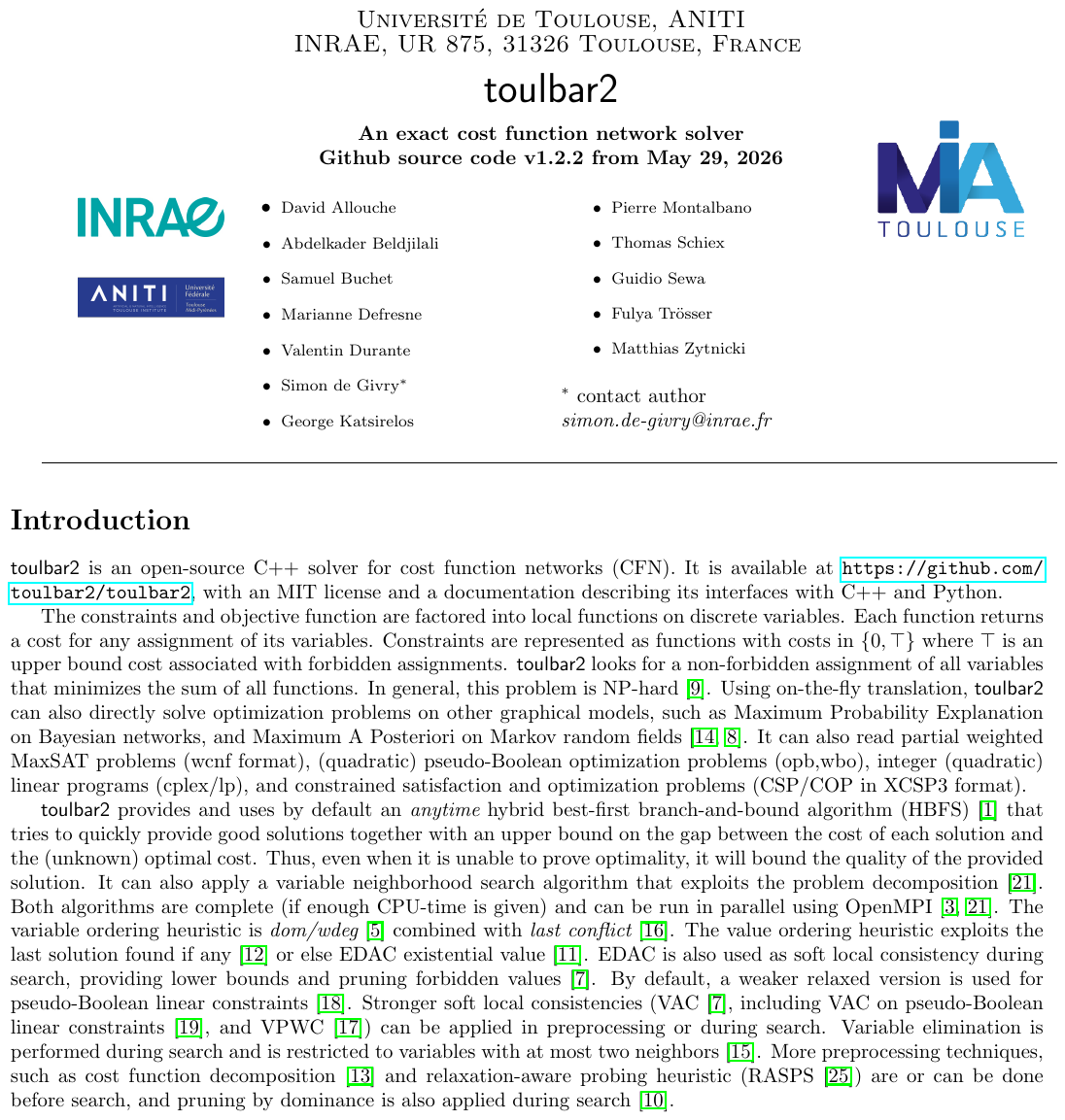}


\chapter{Results}

In this chapter, rankings for the various tracks of the \x3 Competition 2026 are given.
Importantly, remember that you can find all detailed results, including all traces of solvers at \href{https://www.cril.univ-artois.fr/XCSP26/}{https://www.cril.univ-artois.fr/XCSP26/}.

\section{Context}

\bigskip
Remember that the tracks of the competition are given by Table \ref{tab:anysolver} and Table \ref{tab:minisolver}.

\begin{table}[h!]
\begin{center}
\begin{tabular}{cccc} 
\toprule
\textcolor{dred}{\bf Problem} &  \textcolor{dred}{\bf Goal} &  \textcolor{dred}{\bf Exploration} &  \textcolor{dred}{\bf Timeout} \\
\midrule
CSP  & one solution & sequential & 30 minutes \\
COP  & best solution & sequential & 30 minutes \\
Fast COP  & best solution & sequential & 3 minutes \\
// COP  & best solution & parallel & 30 minutes \\
\bottomrule
\end{tabular}
\end{center}
\caption{Standard Tracks. \label{tab:anysolver}}
\end{table}

\begin{table}[h!]
\begin{center}
\begin{tabular}{cccc} 
\toprule
\textcolor{dred}{\bf Problem} &  \textcolor{dred}{\bf Goal} &  \textcolor{dred}{\bf Exploration} &  \textcolor{dred}{\bf Timeout} \\
\midrule
Mini CSP  & one solution & sequential & 30 minutes \\
Mini COP  & best solution & sequential & 30 minutes \\
\bottomrule
\end{tabular}
\end{center}
\caption{Mini-Solver Tracks. \label{tab:minisolver}}
\end{table}

\noindent Also, note that:

 \begin{itemize}
\item The cluster was provided by CRIL and is composed of 10 nodes (bi-processors AMD EPYC 9355, 1 TB de RAM). 
\item For each job, one NUMA zone of a processor AMD EPYC 9355 (8 cores) was allocated.
\item Memory was sotware-limited by runsolver to 64 GB per job.
\item Timeouts were set accordingly to the tracks through the tool \texttt{runsolver}:
  \begin{itemize}
    \item MiniCSP/MiniCOP/CSP/COP : 30 min of CPU time (1h of WC time) 
    \item fast-COP : 3 min of CPU time (1h of WC time) 
    \item parallel-COP : 8 cores, 30 min of WC time (4h of CPU time) 
  \end{itemize}
 \end{itemize}
 

\paragraph{About Scoring.}
The number of points won by a solver $S$ is decided as follows:
\begin{itemize}
\item for CSP, this is the number of times $S$ is able to solve an instance, i.e., to decide the satisfiability of an instance (either exhibiting a solution, or indicating that the instance is unsatisfiable)
\item for COP, this is, roughly speaking, the number of times $S$ gives the best known result, compared to its competitors.
More specifically, for each instance $I$:
\begin{itemize}
\item if $I$ is unsatisfiable, 1 point is won by $S$ if $S$ indicates that the instance $I$ is unsatisfiable, 0 otherwise,
\item if $S$ provides a solution whose bound is less good than another one (found by another competiting solver), 0 point is won by $S$,
\item if $S$ provides an optimal solution, while indicating that it is indeed the optimality, 1 point is won by $S$,
\item if $S$ provides (a solution with) the best found bound among all competitors, this being possibly shared by some other solver(s), while indicating no information about optimality: 1 point is won by $S$ if no other solver proved that this bound was optimal, $0.5$ otherwise.
\end{itemize}
\end{itemize}


\paragraph{Off-competition Solvers.} Some solvers were run while not officially entering the competition: we call them {\em off-competition} solvers.
\ace is one of them because its author (C. Lecoutre) conducted the selection of instances, which is a very strong bias.
Also, when two or more variants (by the same competiting team) of a same solver compete in a same track, only the best one is ranked (and the other ones considered as being off-competition).
To determine which solver variant is the best for a team, we compute the ranking with only the variants of the team competiting between them. 
This is why, for example, in some cases, solvers submitted by the CPMpy team were discarded from the ranking. 

\section{Rankings}

Recall that, concerning ranking, two new rules are used when necessary:
\begin{itemize}
\item In case a team submits the same solver to both the main track and the mini-track for the same problem (CSP or COP), 
the solver will be ranked in the mini-track only if the solver is not one of the three best solvers in the main track.
\item In case several teams submit variations of the same solver to the same track, only
the team who developed the solver and the best other team with that solver will
be ranked (possibly, a second best other team, if the jury thinks that it is relevant)
\end{itemize}

\bigskip
\noindent The algorithm used in practice for establishing the ranking is:
\begin{enumerate}
\item first, off-competition solvers are discarded (this is the case for ACE in 2026)
\item second, only the best variation of the same solver (submitted by the same team) is kept when computing scores (ranking) ; to identify the best variation of a solver, we compute scores while discarding all other solvers (only variataiosn of the solver participate). 
\item third, in mini-tracks, (any variations of) solvers that (whose variations) are ranked 1st, 2nd or 3rd in the corresponding main track are discarded 
\end{enumerate}

\bigskip
Here are the rankings\footnote{The images of medals come from \href{https://www.flaticon.com/free-icons/2nd-place}{Flaticon -- icons created by Md Tanvirul Haque.}} 
     for the 6 tracks.
As PyCSP3\_ortools and  CPMpy\_ortools have obtained close results in some tracks (corresponding to the same ranking), we decided to put them together (ex-aequo in term of ranking) for these tracks. 

\bigskip 

\begin{minipage}{0.4\textwidth}
  \begin{center}
    \hfill
  \begin{tabular}{cp{3.2cm}}
    \multicolumn{2}{l}{{\bf {\Large CSP}}} \\
     \midrule
     \vspace{-0.2cm} &  \\
     \includegraphics[scale=0.075]{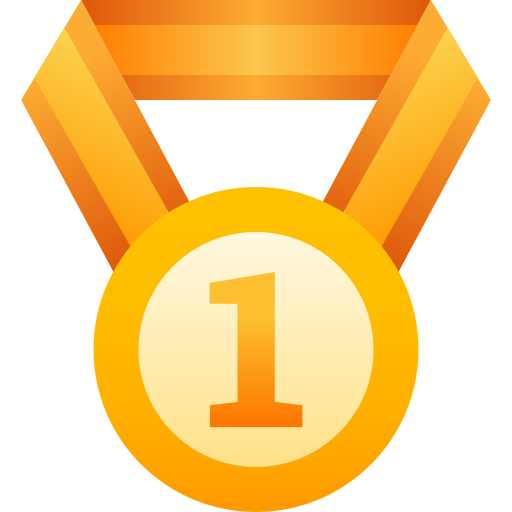} &  \vspace{-0.6cm} {\large kat CPMpy\_ortools} \\
      \midrule
      & \\
      \includegraphics[scale=0.075]{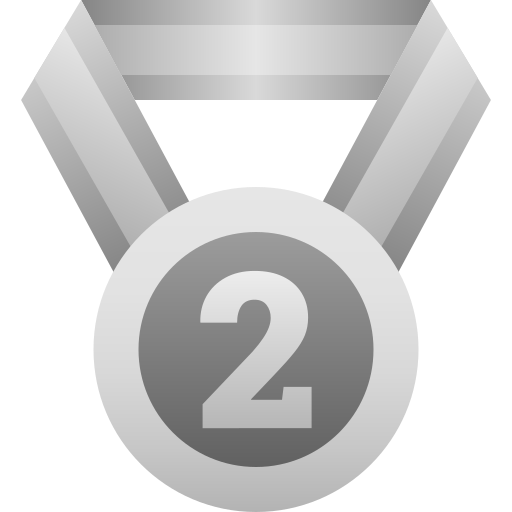} &  \vspace{-0.9cm} \\
       \midrule
      & \\
       \includegraphics[scale=0.075]{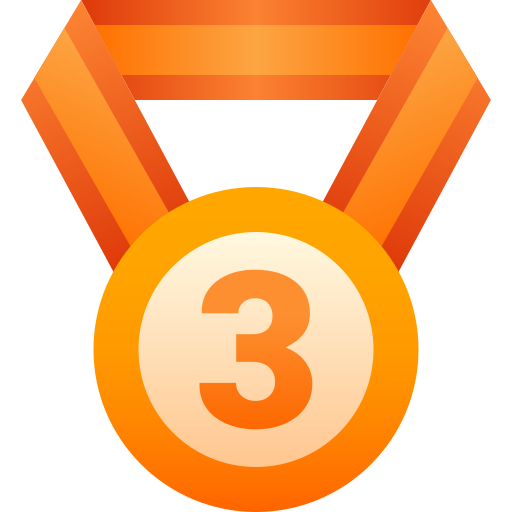} &  \vspace{-0.6cm} {\large Picat} \\
        \midrule
  \end{tabular}
\end{center}
\end{minipage} 
\hfill
\begin{minipage}{0.4\textwidth}
\begin{center}
  \begin{tabular}{cp{3.2cm}}
    \multicolumn{2}{l}{{\bf {\Large COP}}} \\
     \midrule
    \vspace{-0.2cm} &  \\
    \includegraphics[scale=0.075]{gold.png} &  \vspace{-0.6cm}  {\large CoSoCo} \\
     \midrule
      & \\
     \includegraphics[scale=0.075]{silver.png} &  \vspace{-0.6cm} {\large Choco} \\
      \midrule
      & \\
      \includegraphics[scale=0.075]{bronze.png} &  \vspace{-0.6cm} {\large CPMpy\_ortools} \\
       \midrule
  \end{tabular}
\end{center}
\end{minipage}
\hfill

\bigskip\bigskip\bigskip

\begin{minipage}{0.4\textwidth}  
  \begin{center}
    \hfill
  \begin{tabular}{cp{3.2cm}}
    \multicolumn{2}{l}{{\bf {\Large Fast COP}}} \\
     \midrule
    \vspace{-0.2cm} &  \\
    \includegraphics[scale=0.075]{gold.png} &  \vspace{-0.6cm} {\large Choco} \\
     \midrule
       & \\
     \includegraphics[scale=0.075]{silver.png} &  \vspace{-0.6cm} {\large CoSoCo} \\
      \midrule
      & \\
      \includegraphics[scale=0.075]{bronze.png} &  \vspace{-0.6cm} {\large CPMpy\_ortools} \\
       \midrule
  \end{tabular}
\end{center}
\end{minipage}
\hfill 
\begin{minipage}{0.4\textwidth}
\begin{center}
  \begin{tabular}{cp{3.2cm}}
    \multicolumn{2}{l}{{\bf {\Large Parallel COP}}} \\
     \midrule
    \vspace{-0.2cm}   &  \\
    \includegraphics[scale=0.075]{gold.png} &  \vspace{-0.9cm} {\large CoSoCo} \\
     \midrule
      & \\
     \includegraphics[scale=0.075]{silver.png} &  \vspace{-0.6cm} {\large Choco} \\
      \midrule
      & \\
      \includegraphics[scale=0.075]{bronze.png} &  \vspace{-0.6cm} {\large CPMpy\_ortools}  \\
       \midrule
  \end{tabular}
\end{center}
\end{minipage}
\hfill

\bigskip\bigskip\bigskip 

\begin{minipage}{0.4\textwidth}  
  \begin{center}
    \hfill
  \begin{tabular}{cp{3.2cm}}
    \multicolumn{2}{l}{{\bf {\Large Mini CSP}}} \\
    \midrule
    \vspace{-0.2cm} &   \\
    \includegraphics[scale=0.075]{gold.png} &  \vspace{-0.6cm} {\large Qayd} \\
    \midrule
      & \\
    \includegraphics[scale=0.075]{silver.png} &  \vspace{-0.6cm} {\large Sat4j} \\
    \midrule
      & \\
    \includegraphics[scale=0.075]{bronze.png} &  \vspace{-0.6cm} {\large Tempo} \\
    \midrule
  \end{tabular}
\end{center}
\end{minipage}
\hfill 
\begin{minipage}{0.4\textwidth}
\begin{center}
  \begin{tabular}{cp{3.2cm}}
     \multicolumn{2}{l}{{\bf {\Large Mini COP}}} \\
     \midrule
    \vspace{-0.2cm} &  \\
    \includegraphics[scale=0.075]{gold.png} &  \vspace{-0.6cm} {\large Qayd} \\
      \midrule
      & \\
      \includegraphics[scale=0.075]{silver.png} &  \vspace{-0.6cm} {\large Tempo} \\
        \midrule
      & \\
        \includegraphics[scale=0.075]{bronze.png} &  \vspace{-0.6cm} {\large toulbar2} \\
          \midrule
  \end{tabular}
\end{center}
\end{minipage}
\hfill

\pagebreak
\chapter*{Ackowledgments}

This work was supported by state funding managed by the French National Research Agency (ANR) under the France 2030 program, as well as by the European Union through the NextGenerationEU program (reference 22-EXES-0009).


\end{document}